\documentclass[a4paper]{cas-sc}

\usepackage[numbers]{natbib}
\usepackage{subcaption}  % in preamble
\usepackage[linesnumbered, ruled]{algorithm2e}
\usepackage{graphicx} % Required for inserting images
\usepackage{amsmath}
\usepackage{float}
\usepackage[shortlabels]{enumitem}

\usepackage{tabularx}
\usepackage{array}    % 若你还没用
\newcolumntype{M}[1]{>{\centering\arraybackslash}m{#1}} % 如果你之前有 M 定义可以忽略

\usepackage{varwidth}
\DeclareUnicodeCharacter{FF1A}{:}

\newcommand{\State}{}

\renewcommand{\Return}[1]{\textbf{return }#1\;}
\newcommand{\Endif}{}
\newcommand{\EndFor}{}
\let\origleftarrow\leftarrow
\renewcommand{\leftarrow}{\ensuremath{\origleftarrow}}
\let\origcup\cup
\renewcommand{\cup}{\ensuremath{\origcup}}
\let\origcap\cap
\renewcommand{\cap}{\ensuremath{\origcap}}
\let\origemptyset\emptyset
\renewcommand{\emptyset}{\ensuremath{\origemptyset}}
\let\origtimes\times
\renewcommand{\times}{\ensuremath{\origtimes}}
\let\origleq\leq
\renewcommand{\leq}{\ensuremath{\origleq}}
\let\origle\le
\renewcommand{\le}{\ensuremath{\origle}}
\let\origgeq\geq
\renewcommand{\geq}{\ensuremath{\origgeq}}
\let\origeq\neq
\renewcommand{\neq}{\ensuremath{\origeq}}
\let\origin\in
\renewcommand{\in}{\ensuremath{\origin}}
\let\origoverline\overline
\renewcommand{\overline}[1]{\ensuremath{\origoverline{#1}}}

\def\tsc#1{\csdef{#1}{\textsc{\lowercase{#1}}\xspace}}
\tsc{WGM}
\tsc{QE}
\begin{document}
\let\WriteBookmarks\relax
\def\floatpagepagefraction{1}
\def\textpagefraction{.001}

\shorttitle{}    

\shortauthors{}  

\title [mode = title]{A Large-Scale Sparse Multiobjective Optimization Algorithm Based on Optimal Performance Scores}  

\author[1]{Jia-Lin Mai}
\author[1]{Min-Rong Chen}\cormark[1]\ead{mrongchen@126.com}
\author[2]{Guo-Qiang Zeng}
\author[3]{Xiang Liu}\cormark[2]\ead{liuxiang@dgut.edu.cn}
\author[4]{Jian Weng}

\affiliation[1]{organization={School of Computer Science, South China Normal University},
                city={Guangzhou},
                postcode={510631},
                country={China}}

\affiliation[2]{organization={National-local Joint Engineering Laboratory for Digitalize Eletrical Design Technology, Wenzhou University},
                city={Wenzhou},
                postcode={325035},
                country={China}}

\affiliation[3]{organization={Institute of Science and Technology Innovation, Dongguan University of Technology},
                city={Dongguan},
                postcode={523808},
                country={China}}

\affiliation[4]{organization={College of Cyber Security, Guangzhou University},
                city={Guangzhou},
                postcode={510006},
                country={China}}

\cortext[cor1]{Corresponding author: Min-Rong Chen}
\cortext[cor2]{Corresponding author: Xiang Liu}

\begin{abstract}
Large-scale sparse multiobjective optimization problems (LSSMOPs) involve a large number of decision variables and Pareto optimal solutions with only a few nonzero variables. However, as the number of decision variables grows, it becomes increasingly challenging to accurately identify the nonzero variables, and optimization performance is adversely affected. To address these issues, this paper proposes an evolutionary algorithm for LSSMOPs. Specifically, we propose a new initialization method capable of generating scores that accurately reflect the importance of variables, and an initial mask vector template that can locate nonzero variables. This leads to the generation of a high-quality initial population. Additionally, this paper introduces a new strategy to calculate the mutation probability for each variable and a novel optimization for real variables based on the Pareto-guided normal distribution, enabling the population to avoid being trapped in local optima and quickly converge to the global optimum. Experimental results from eight benchmark problems and three real-world applications demonstrate that the proposed algorithm achieves superior performance compared with state-of-the-art algorithms.
\end{abstract}



% Keywords

\maketitle

% Main text
\section{Introduction}\label{}
Multiobjective optimization plays a central role in many fields~\cite{coello2006evolutionary}, such as machine learning~\cite{machinelearning, zhou2011multiobjective}, data mining~\cite{tang2012incorporating}, and financial decision making~\cite{economics}, aiming to optimize multiple (often conflicting) performance criteria simultaneously and to obtain the Pareto front (PF). Pareto optimal solutions of sparse multiobjective optimization problems (SMOPs) contain only a few nonzero variables. SMOPs have direct engineering relevance in a variety of applications~\cite{qi2024enhancing}, including feature selection~\cite{xue2013particle}, where one seeks a compact subset of informative features to improve model interpretability and predictive performance; sparse signal reconstruction~\cite{li2018preference}, which aims to recover an accurate signal from limited or noisy measurements by exploiting its inherent sparsity; and sparse neural network training~\cite{fieldsend2005pareto}, which enforces weight sparsity so that most network parameters become zero and the resulting model is smaller to store. In SMOPs, we seek solutions that are both close to the PF and as sparse as possible, to satisfy practical constraints like interpretability, storage, and computational cost.\\ 
\indent With the rapid growth of data scale and system complexity, large-scale sparse multiobjective optimization algorithms (LSSMOOAs) have become increasingly common: decision-variable dimensionality can be extremely high, but in many practical problems, only a small subset of variables dominantly affects objectives (i.e., solutions are sparse). This sparsity is both a challenge and an opportunity -- exploiting sparse structure can dramatically reduce the search space and evaluation cost, while ignoring it often leads conventional algorithms to suffer from the curse of dimensionality or poor efficiency~\cite{tan2021multi}.\\
\indent Conventional multiobjective evolutionary algorithms (MOEAs) perform well on small-scale benchmark problems but encounter three major bottlenecks on large-scale SMOPs~\cite{shao2025evolutionary, yang2025sparse,huang2025enhanced}. First, random variation and standard crossover make it difficult to find key variable combinations in high-dimensional spaces, leading to low search efficiency~\cite{zhou2024highdim,hua2021irregularpf}. Second, function evaluations are often expensive, and thus repeated evaluations of high-dimensional solutions become prohibitive~\cite{7544478}. Third, maintaining both convergence and diversity becomes harder under sparsity constraints~\cite{jiang2025heterogeneous, guo2025constrained}: algorithms may converge to dense or local optimal solutions~\cite{wang2025cmoea}. Recent advances -- such as sparsity-aware operators~\cite{A_CO_E}, variable selection heuristics~\cite{jiang2023two}, neural network model based dimensionality reduction method~\cite{ye2022solving,10477568}, and multi-stage knowledge guidance~\cite{ye2024mskea} -- have partially addressed these problems, but important limitations remain: many methods rely on static or heuristic variable selection rules that fail to track changes in variable importance during the searc~\cite{wang2025evolution, liu2021variable}; others lack effective coordination between neural network models and population structure, which limits scalability and robustness in real large-scale tasks.~\cite{li2025surrogate, liang2024multi, santos2023neuroevolution}\\
% ====================================================
% ====================================================
\indent To address these challenges, this paper proposes an algorithm that can efficiently optimize binary vectors and real vectors. Specifically, the main contributions are as follows:\\
\begin{enumerate}[(\arabic*)]
	\item In this paper, we propose a new initialization framework that can be easily embedded into other algorithms. The scores of the variables are calculated based on their performance in all intervals of the decision space, which can more accurately reflect the importance of the variables. Based on these scores, an initial mask vector template can be generated, which guides the initial population to locate the nonzero variables. The values of the real variables are sampled in the appropriate intervals according to their performance in different intervals. Therefore, the initialization method can generate a high-quality initial population that accurately identifies the nonzero variables.
    
	\item We propose an adaptive mutation probability for each real variable and a Pareto-guided resampling strategy based on the normal distribution. The mutation scheme sets the mutation probability for each real variable based on its behavior across all intervals, assigning higher probabilities to variables that are more prone to becoming trapped in local optima. The Pareto-guided resampling strategy facilitates the generation of real vectors of Pareto optimal solutions.

	\item The proposed large-scale sparse multiobjective optimization algorithm based on optimal performance scores, abbreviated as SMOEA-OPS, is used to compare with the state-of-the-art on eight benchmark problems and three real-world problems. The experiments demonstrate that SMOEA-OPS achieves significantly better performance than competing algorithms.
\end{enumerate}

\indent The rest of this paper is organized as follows. In Section II, some basic concepts of sparse multiobjective optimization problems, existing LSSMOOAs, and the motivation of this work are given. In Section III, the proposed algorithm is described in detail. In Section IV, the experimental results on benchmark and real-world SMOPs are presented and analyzed. Finally, the conclusions are described in Section V.
\section{Related work}
\subsection{Sparse multiobjective optimization}
The unconstrained multiobjective optimization problem can be defined mathematically as follows:
\begin{equation}
	\min_{\mathbf{x} \in \mathbb{R}^D} \quad 
	\mathbf{F}(\mathbf{x}) = \left( f_1(\mathbf{x}),\, f_2(\mathbf{x}),\, \dots,\, f_m(\mathbf{x}) \right)
\end{equation}
where $\mathbf{x} = (x_1, x_2, \dots, x_D)^\mathrm{T}$ denotes the decision vector in a $D$-dimensional search space, and $\mathbf{F}(\mathbf{x})$ represents the objective vector consisting of $m$ conflicting objective functions $f_i(\mathbf{x})$, $i = 1, \dots, m$. The goal of multiobjective optimization problems is to find a set of Pareto optimal solutions, for which no objective can be improved without degrading at least one other objective~\cite{marler2004survey}. The concept of Pareto dominance is used to compare solutions. Given two decision vectors $\mathbf{x}^a$ and $\mathbf{x}^b$, we say that $\mathbf{x}^a$ \emph{dominates} $\mathbf{x}^b$, denoted by $\mathbf{x}^a \prec \mathbf{x}^b$, 
if and only if the following two conditions hold:
\begin{enumerate}
	\item $f_i(\mathbf{x}^a) \le f_i(\mathbf{x}^b)$ for all $i \in \{1, \dots, m\}$,
	\item $f_j(\mathbf{x}^a) < f_j(\mathbf{x}^b)$ for at least one index $j \in \{1, \dots, m\}$.
\end{enumerate}
a solution is called Pareto optimal if it is not dominated by any other solution in the search space. 
The set of all Pareto optimal solutions is known as the Pareto set (PS).\\ 
\indent Sparse multiobjective optimization, on the other hand, means that only a few decision variables in the decision vectors of the Pareto optimal solutions are nonzero, i.e., the number of meaningful variables is much smaller than the total number of decision variables \(D\). When dealing with LSSMOPs (\(D\)\(\ge\)100), conventional MOEAs often suffer from prohibitive computational costs due to the curse of dimensionality and the difficulty of effectively identifying the influential variables~\cite{tian2021evolutionary}. To address these challenges, a variety of sparse multiobjective optimization algorithms have been developed.\\
\subsection{Initialization for solving LSSMOPs}
Existing algorithms for tackling sparse problems typically use one of three initialization strategies~\cite{wang2024sparse}: random sampling, Latin hypercube sampling~\cite{mckay2000comparison}, or the sparse population approach. For example, S-ECSO~\cite{wang2022enhanced} and PM-MOEA~\cite{tian2022pattern} use random initialization, i.e., the initial $N$ individuals are generated at random positions in the decision space. There are also a few algorithms initialized by the Latin hypercube sampling method, such as MOEA/PSL~\cite{ye2022solving}.\\
\indent However, the majority of LSSMOOAs employ the sparse population approach, which was originally proposed by the first sparse MOEA -- SparseEA~\cite{tian2020evolutionary}. SparseEA first introduced representing an individual $\mathbf{x}$ using a real vector \(dec\) and a binary vector \(mask\), i.e., a two-layer encoding scheme, as given by the following formula:
\begin{equation}
	\mathbf{x} = \big( dec_1\times mask_1,\; \dots,\; dec_D\times mask_D \big)
\end{equation}
where \(D\) denotes the dimension of the solutions. In SparseEA, the initialization begins by constructing \(D\) sparse solutions through the two-layer encoding strategy. In each of these solutions, only one decision variable is set to a nonzero value, while all others are fixed at zero. These \(D\) solutions are evaluated and ranked according to Pareto dominance, and a score is assigned to each decision variable based on the non-dominated front number of its associated single-variable solution. From these scores, we can generate a binary mask matrix that determines the active variables in each individual. This binary mask matrix is then combined element-wise with a randomly generated real-valued matrix to form the initial population.\\ 
\indent However, the scores produced by SparseEA can be unreliable. Recent methods such as MGCEA~\cite{MGCEA} and KLEA~\cite{shao2025knowledge} address this by splitting each domain of decision variables into multiple equal subintervals and sampling within each interval. For each variable, MGCEA computes the score by aggregating its non-dominated front number across all sampled intervals. By evaluating a variable’s performance throughout its range rather than at random ranges, this multi-interval sampling yields more robust and accurate scores for constructing a sparse initial population.
\subsection{Real vectors optimization for LSSMOPs}
Most LSSMOPs now adopt a two-layer encoding that represents a solution with a hybrid of a real vector and a binary vector. While most LSSMOOAs focus on handling sparsity via mask encodings without directly optimizing the real variables, a few algorithms specifically address convergence on the real vector component. For example, SparseEA2~\cite{SparseEA2} enhances the mask–real interplay via variable grouping to ensure nonzero variables are more thoroughly optimized. MOEA/PSL achieves dimensionality reduction through autoencoder-based subspaces, where real vectors are optimized in a learned low-dimensional domain. MSKEA~\cite{ye2024mskea}, S-ECSO, and PM-MOEA incorporate dynamic sparse knowledge or convex operators to guide the real vector toward Pareto optimal. Furthermore, rank-1 approximation-based EAs~\cite{chen2023rank1} decompose the real vector into subspaces via tensor decomposition and optimize them alternately, enabling more efficient convergence.
\subsection{Motivation of this work}
\indent The scoring mechanism used by SparseEA fails to provide a reliable measure of variable importance, since it does not consider the full distribution of each variable’s values within the decision space. MGCEA computes each variable’s score by summing its non-dominated front numbers across all intervals. However, this score calculation method also tends to ignore some variables that have potential. As shown in Fig. 1, the non-dominated front number of the first variable in the third interval is very small, i.e., this variable is very promising. But the sum of the non-dominated front numbers in all intervals makes the score of this variable very large, which can easily result in overlooking this variable, especially in some complex landscapes. 

To enable subsequent individuals to evolve based on more accurate scores, this paper considers a new method for computing scores. In this paper, each variable’s score is determined by the lowest non-dominated front number it attains over the decision space. The non-dominated front numbers of each variable across all intervals are recorded in the matrix \(IntervalFronts\). This matrix can also guide which interval each real variable should be sampled from, thereby producing high-quality initial solutions.\\
\indent Since only a very small number of LSSMOOAs consider the optimization of real vectors, this paper proposes a Pareto-guided resampling method based on the normal distribution for optimizing real vectors. We divide the evolution into two stages according to the number of decision variables. In stage one, we preserve the diversity of the real vectors; in stage two, we intensify resampling to steer solutions toward the Pareto set.

\begin{figure}
	\centering
	\includegraphics[width=0.5\textwidth]{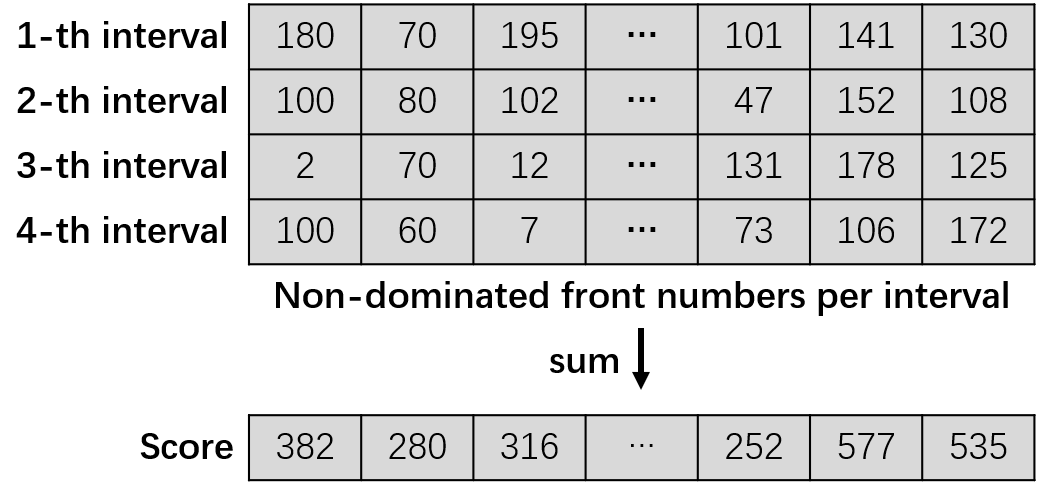} % 不需要.svg后缀
	\caption{Illustrative example of per-variable score calculation in MGCEA.}
	\label{fig:related1}
\end{figure}
\section{The proposed SMOEA-OPS}
	
\subsection{The framework of SMOEA-OPS}
\begin{algorithm}[H] %算法环境
		\caption{Framework of the Proposed SMOEA-OPS}\label{alg:alg1}
		\SetKwData{Left}{left}\SetKwData{This}
		{this}\SetKwData{Up}{up}
		\SetKwFunction{Union}{Union}
		\SetKwFunction{FindCompress}
		{FindCompress} \SetKwInOut{Input}{Input}
		\SetKwInOut{Output}{Output}
		
		\Input{\(N\) (population size), \(FE_{max}\) (maximum number of evaluations), \(NVal\) (number of intervals), \(SVal\) (number of samplings per interval)}
		\Output{\(P\) (final population)}
		\State \([P,Score,flag]\leftarrow~ Initialization(NVal,\ SVal,\ N);\)\tcp*[h]{Algorithm 2} \\
		\While{\(termination\ criterion\ not\ fulfilled\)}{
			
			\State \(P'\) \leftarrow \textup{Select } \(2N\) \textup{parents from} \(P\) \textup{via the mating selection strategy of SPEA2;}
			
			\If{$rand()$ $\leq$ $FE/FE_{max}$} {
				\State \(Sparsity \leftarrow \textup{Caculate the Sparsity of each variable in Pareto binary vectors by Eq.(3)}\)\;
				\State \(Score \leftarrow \textup{Update }Score\textup{ by Eq.(4)}\)\;
			}\Endif
			\State \([\mu, \sigma] \leftarrow\) \textup{Calculate the mean and the standard deviation of each real variable across the current Pareto optimal solutions by Eq.(5) and Eq.(6)}\;
		      \State \(Q'\leftarrow Variation(P',Score,FE/FE_{max},flag);\)
			
			\State \(Q \leftarrow PGResample(Q', \mu, \sigma, FE/FE_{max});\)
			
			\State \(P \leftarrow \textup{Select } N \textup{solutions from } 
				P~\cup\ Q \textup{ via the environmental selection strategy of SPEA2;}\)  
		}
		\Return{\(P\)}
		
	\end{algorithm}
\indent The SMOEA-OPS framework is shown in Algorithm 1. First, we initialize to obtain the population \(P\), the vector \(Score\), and a boolean flag indicating whether the \(initMask\) is informative. In the main loop, \(2N\) parents are selected in \(P\) using the strategy of SPEA2~\cite{zitzler2001spea2} to generate offspring. The vector \(Score\) is used to measure the importance of each decision variable across the solution set; smaller values indicate higher importance. To avoid premature convergence to local optima caused by always using high-quality solutions from the current population to guide offspring, we use the ratio of consumed evaluations as the probability to update \(Score\). Before updating \(Score\), update the vector \(Sparsity\) as follows:
	\begin{equation}
	Sparsity_i = \frac{1}{|PS|}\sum_{p\in PS} pb_i
	\label{eq:sparsity}
	\end{equation}
    where \(Sparsity_i\) is the sparsity of the \(i\)-th variable, \(pb_i\) is the \(i\)-th variable of the binary vector of solution \(p\) in Pareto optimal solutions. Then the score of the \(i\)-th variable of the current generation is updated:
	
	\begin{equation}
        \small
	   	Score_i = Score_i' + \lambda(Sparsity_{max}-Sparsity_i)
		\label{eq:score}
	\end{equation}
    where \(Score_i'\) refers to the score of \(i\)-th variable before it is updated, \(Sparsity_{max}\) denotes the maximum sparsity among all variables in binary vectors, and \(\lambda\) denotes the ratio of consumed evaluations, i.e., $FE/FE_{max}$. $FE$ and $FE_{max}$ represent the current and the maximum number of evaluations, respectively. As the number of function evaluations increases, the scores of zero variables increase much faster than those of nonzero variables, causing the nonzero variables to become increasingly distinguishable. The probability of updating the \(Score\) vector becomes larger as it gets further into the evolutionary process. This enables learning towards high-quality Pareto optimal solutions and achieves the effect of exploration in the early stage and exploitation in the later stage.\\
    \indent Then we compute $\mu$ and $\sigma$ (the mean and standard deviation of each real variable across the current Pareto set) and use them to make offspring learn from the Pareto set. The $i$-th entry of $\mu$ is calculated as:
    \begin{equation}
    \mu_i = \frac{1}{|PS|} \sum_{p \in PS} pr_i
    \end{equation}
    where $pr_i$ denotes the $i$-th real variable of solution $p$ in the Pareto set. The $i$-th element of $\sigma$ is computed as:
    \begin{equation}
    \sigma_i = \sqrt{\frac{1}{|PS|}\sum_{p\in PS}\big(pr_i - \mu_i\big)^2}
    \label{eq:sigma}
    \end{equation}
	\indent Then, a population \(Q\) is generated through variation and resampling. Afterward, population \(Q\) is combined with \(P\). \(N\) solutions are selected from the combined population to survive to the next generation through the environmental selection strategy of SPEA2.\\
	
	% =============================算法2=====================
	\subsection{The proposed initialization approach}
    \begin{algorithm}[H]
		\caption{\(Initialization(NVal,\ SVal,\ N)\)}\label{alg:alg1}
		\SetKwData{Left}{left}\SetKwData{This}
		{this}\SetKwData{Up}{up}
		\SetKwFunction{Union}{Union}
		\SetKwFunction{FindCompress}
		{FindCompress} \SetKwInOut{Input}{Input}
		\SetKwInOut{Output}{Output}
		\Input{\(NVal\) (number of intervals), \(SVal\) (number of samplings per interval), \(N\) (population size)}
		\Output{ \(P\) (initial population), \(Score\) (scores of decision variables), \(flag\) (flag indicating whether \(initMask\) is informative)}
		
		\State \([Score,IntervalFronts]\leftarrow getScore(NVal,SVal,N);\)\tcp*[h]{Algorithm 3} \\
		\State \(D \leftarrow \textup{Number of decision variables}\)\;    
		\State \(T \leftarrow \emptyset\)\;
		\State \(flag \leftarrow 0\)\;

		\For{$i = 2$ to $(lnD)$} {
            \State \(minGroup \leftarrow \textup{the cluster with the smallest mean score after \(K\)-means clustering into \(i\) groups}\)\;
            \State \(maskTpl \leftarrow 1 \times D \textup{ vector of zeros}\)\;
			\For{$j \in minGroup$}{
				\State \(maskTpl_j \leftarrow \textup{1;}\)
			}\EndFor
			\State \(Q\leftarrow~ generateSolutions(N/((lnD)-1),\ IntervalFronts,\ maskTpl);\)\tcp*[h]{Algorithm~4} \\
			\State \(T \leftarrow T\ \cup \ Q\)\;
		}\EndFor
		
		\State \(initMask\)~\leftarrow~\textup{\(getMask\)(\(T,Score\));} \tcp*[h]{Algorithm 5} \\
		\If{\(any\ element\ of\ initMask\ is\ nonzero\)}{
			\State \(P \leftarrow generateSolutions(N,\ IntervalFronts,\ initMask);\)
			\tcp*[h]{Algorithm 4}\\
		}\Else{
			\State \(flag \leftarrow 1\)\;
			\(P\) \leftarrow \(T\)\;
		}\Endif
		
		\Return{\([P,\ Score,\ flag]\)}
		
	\end{algorithm}
    
	The initialization framework is shown in Algorithm 2. The vector \(Score\) and the matrix \(IntervalFronts\) are first calculated using Algorithm 3. The \(flag\) indicates whether \(initMask\) is informative. The mask template \(initMask\) is used to guide the generation of the binary vectors of the initial population. If all elements of \(initMask\) are zero, it provides little guidance, and \(flag\) is set to 1. The generation of population T, from which \(initMask\) is derived, is described in Lines 5–13 of Algorithm 2 (see Fig. 2 for an example).\\
    \indent First, $(ln D)-1$ iterations are carried out, where $D$ denotes the number of decision variables, and the loop index $i$ starts at $2$. In each iteration, $K$-means is first applied to partition all variables into $i$ groups based on their scores. Variables in the group with the smallest mean score are set to $1$, while variables in the remaining groups are set to $0$. From each iteration, $N / ((lnD)-1)$ identical mask vectors are produced, so that after all iterations a total of $N$ mask vectors are obtained. Meanwhile, $N$ real vectors are generated based on $IntervalFronts$. The calculation of \(IntervalFronts\) and the generation of solutions will be described in Algorithm 3 and Algorithm 4, respectively. Subsequently, the $N$ mask vectors and the $N$ real vectors are combined to form a population $T$ of size $N$, which is then utilized to generate an \(initMask\) for the initialization of the population’s binary vectors. The population $T$ not only maintains sparsity but also activates the most promising variables.
    \begin{algorithm}
		\caption{\(getScore(NVal,\ SVal,\ N)\)}\label{alg:alg1}
		\SetKwData{Left}{left}\SetKwData{This}
		{this}\SetKwData{Up}{up}
		\SetKwFunction{Union}{Union}
		\SetKwFunction{FindCompress}
		{FindCompress} \SetKwInOut{Input}{Input}
		\SetKwInOut{Output}{Output}
		
		\Input{\(NVal\) (number of intervals), \(SVal\) (number of samplings per interval), \(N\) (population size)}
		\Output{\(Score\) (scores of decision variables), \(IntervalFronts\) (the matrix of non-dominated front numbers for variables across intervals)}
		
		\State $D \leftarrow$ \textup{Number of decision variables}\;
		\State $IntervalFronts \leftarrow$ \textup{a zero matrix with $NVal$ rows and $D$ columns}\;
		
		\For{$m = 1$ to $NVal$}{
			\State $Front \leftarrow$ \textup{$1 \times D$ vector of zeros}\;
			\For{$n = 1$ to $SVal$}{
				\State $Dec \leftarrow$ \textup{A $D \times D$ real-valued matrix, where the $D$ entries in column $c$ are generated by Eq.(7)}\;
				\State $Mask \leftarrow$ \textup{$D \times D$ identity mask}\;
				\State $Q \leftarrow$ \textup{A population of $D$ solutions, where the $r$-th solution is generated from the $r$-th rows of $Dec$ and $Mask$ according to Eq.(2)}\;
				\State \(\textup{Apply non-dominated sorting to all solutions in population $Q$}\)\;
				\For{$i = 1\ to\ D$}{
					\State $rankND_i \leftarrow$ \textup{Non-dominated front number of the $i$-th solution in $Q$}\;
                    \State $Front_i \leftarrow Front_i + rankND_i;$
				}
				\EndFor
			}
			\EndFor
			\State \textup{Replace the elements in the $m$-th row of $IntervalFronts$ with those of vector $Front$;}
		}\EndFor
		
		\State $Score \leftarrow$ \textup{$1 \times D$ vector of ones}\;
		\For{$i = 1$ to $NVal$} {
			\State $maxFro \leftarrow$ \textup{maximum value in the $i$-th row of $IntervalFronts$}\;
			\For{$j = 1$ to $D$} {
				\State $Score_j \leftarrow min(Score_j,\ IntervalFronts_{i,j}/maxFro)$\;
			}
		}\EndFor
		\Return{\([Score,\ IntervalFronts]\)}
		
	\end{algorithm}
    
\begin{figure}[]
		\centering
        \includegraphics[width=1\textwidth]{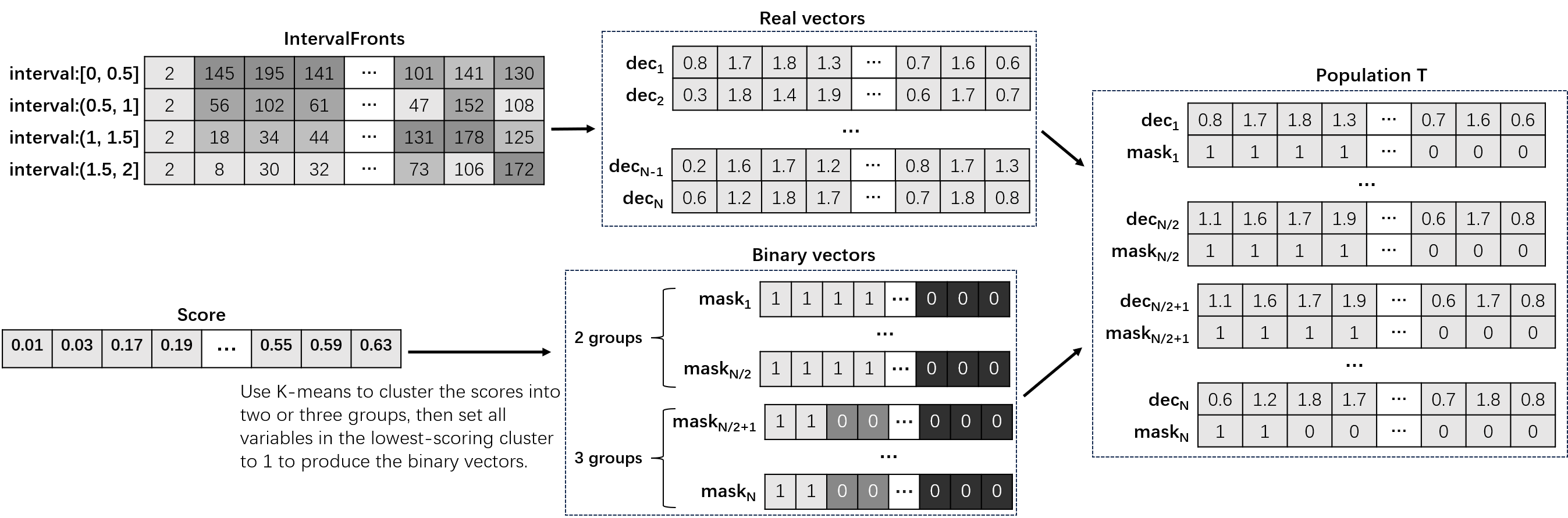}
		\caption{Illustrative example of generating a population $T$. Assume real variables range over [0,2], $lnD$=3, and \(NVal\)=4. In the binary vectors, darker shading indicates higher variable scores within a group. In each column of the \(IntervalFronts\) matrix, lighter shading corresponds to a smaller non-dominated front number for the associated variable in that interval — in other words, lighter entries make the variable more likely to take a value from that interval.}
		\label{fig:svg}
	\end{figure}
    \begin{figure}
		\centering
		\includegraphics[width=1\textwidth]{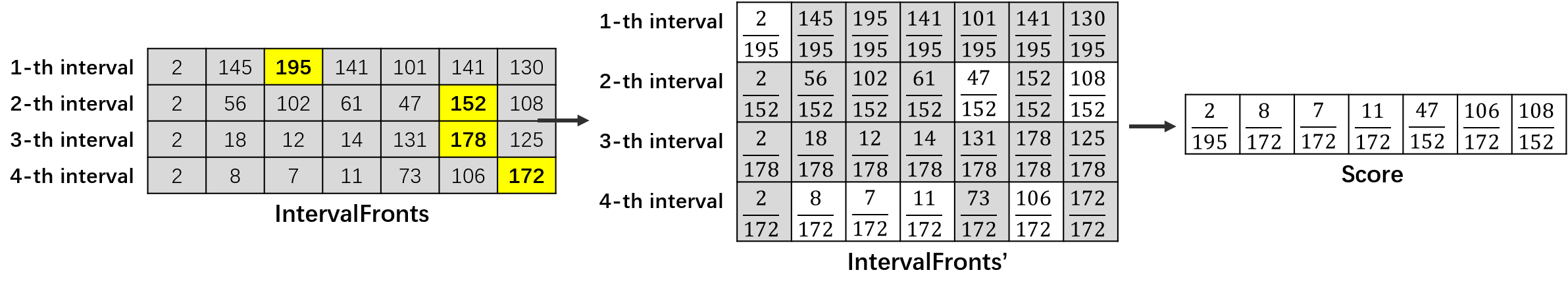} % 不需要.svg后缀
		\caption{Illustrative example of the proposed SMOEA-OPS's score calculation method. In the matrix \(IntervalFronts\), the yellow entries mark the maximum values in their respective rows, while in matrix \(IntervalFronts'\), the white entries denote the minimum values in their respective columns.}
		\label{fig:svg}
	\end{figure}
    
    \indent To obtain mask vectors that more accurately identify nonzero variables, an initial mask template $initMask$ is computed from the population $T$ and the variable-score vector $Score$; this template is used to guide the generation of initial mask vectors. If all elements of the computed $initMask$ are zero, $T$ is used directly as the initial population; otherwise, $initMask$ and $IntervalFronts$ are used separately to generate $N$ mask vectors and $N$ real vectors for the initial solutions.\\
    \indent The maximum number of groups is set to $ln D$ because it adapts to $D$: as the number of decision variables $D$ increases, partitioning becomes finer. Moreover, $ln D$ grows much more slowly than $D$, keeping $ln D$ within a reasonable range as $D$ varies and thereby reducing computational cost.

	% ============================== 算法3 =========================
	% ============================== 算法3 =========================
	
	\indent Algorithm 3 details the method for calculating \(Score\) during the initialization phase. The \(Front\) calculation is similar to the multi-interval sampling~\cite{MGCEA} proposed by MGCEA, which allows the total non-dominated front numbers of each variable across the entire interval to be taken into account when calculating $Score$. The range of the real value of the \(c\)-th variable in the \(m\)-th iteration is as follows:
    \begin{equation}
	\begin{split}
		Dec_{\cdot c} \in & \left[ lower_c + \frac{(upper_c - lower_c)\times(m-1)}{NVal}, \right. \\
		& \left. lower_c + \frac{(upper_c - lower_c)\times m}{NVal} \right]
	\end{split}
	\label{eq:sparsity}
    \end{equation}
    where \(lower_c\) and \(upper_c\) are the lower and upper bounds of the \(c\)-th real variable, respectively.\\
    \indent However, if we simply add up the non-dominated front numbers of all intervals to calculate the scores may overlook potential variables. When certain variables perform exceptionally well in one interval but poorly in others, the score obtained by adding up the non-dominated front numbers of all intervals may be high, resulting in the variable being eliminated. To solve this problem, the non-dominated front numbers of each variable in the \(NVal\) intervals should be retained as \(IntervalFronts\). For example, 
    $IntervalFronts_{i,j}$ denotes the sum, over $SVal$ runs of non-dominated sorting, of the non-dominated front numbers of solutions in interval $i$ drawn from the solution sets where only variable $j$ is nonzero. To explain more clearly, 
    Fig. 3 shows an example of the initialization of \(Score\). First, the matrix \(IntervalFronts'\) is obtained by dividing each element of the \(IntervalFronts\) matrix by the maximum element value in its corresponding row. Then, the minimum value in each column of $IntervalFronts'$, corresponding to that variable's score at its optimal performance, is used to form the vector $Score$. The lower a variable’s score, the more important that variable tends to be.\\
	\indent The population generated during initialization is shown in Algorithm 4. Higher-quality solutions can be generated in the initialization by using \(IntervalFronts\) to guide the generation of real vectors, and \(initMask\) to guide the generation of binary vectors. When generating real vectors, each variable should be sampled with high probability in the interval with a low non-dominated front number. So the probability that the $j$-th variable is sampled from the $intv$-th interval is given as follows:
	\begin{equation}
		\large
		intvProb_{j,intv} = \frac{\frac{1}{IntervalFronts_{intv,j}}}{\sum_{i=1}^{NVal}\frac{1}{IntervalFronts_{i,j}}}
	\end{equation}
    where \(NVal\) is the number of intervals. Taking \(IntervalFronts\) of Fig. 3 as an example, the probability that the 2-th real variable is sampled in the 3-th interval is $\frac{1}{18}/(\frac{1}{8}+\frac{1}{18}+\frac{1}{56}+\frac{1}{145})$. In this case, each real variable has a greater chance of generating real values in the range where it performs well, thereby obtaining good real vectors. For binary vectors, the value of each variable in the mask template represents the probability that the corresponding variable should be set to 1 when generating binary vectors. \\
    \begin{algorithm}[H]
		\caption{\(generateSolutions(N,\ IntervalFronts,\ maskTpl)\)}\label{alg:alg1}
		\SetKwData{Left}{left}\SetKwData{This}
		{this}\SetKwData{Up}{up}
		\SetKwFunction{Union}{Union}
		\SetKwFunction{FindCompress}
		{FindCompress} \SetKwInOut{Input}{Input}
		\SetKwInOut{Output}{Output}
		\Input{\(N\) (population size), \(IntervalFronts\) (non-dominated front numbers of all intervals), \(maskTpl\) (the mask template)}
		\Output{ \(P\) (a population of \(N\) individuals generated based on \(IntervalFronts\) and \(maskTpl\))}
		
		\State $P \leftarrow \emptyset$\;
		\State $Dec \leftarrow \emptyset$\;
		\State $Mask \leftarrow \emptyset$\;
		\State $D \leftarrow$ \textup{Number of decision variables}\;
		\tcp*[h]{Generate \(N\) solutions}\\
		\For{$i = 1$ to $N$} {
			\State $TDec \leftarrow$ \textup{$1 \times D$ vector of zeros}\;
			\State $TMask \leftarrow$ \textup{$1 \times D$ vector of zeros}\;
			\For{$j = 1$ to $D$} {
				\If{$rand()$ \le $maskTpl_j$}{
					\State $TMask_j \leftarrow 1$\;
                    \State \textup{$intv \leftarrow $ Select an interval with probability calculated by Eq.(8)}\;
                    \State \textup{$TDec_j \leftarrow$ Generate a random value within the $intv$-th interval;}
				}\Endif
			}\EndFor
			\State $Dec \leftarrow [Dec;\ TDec]$\;
			\State $Mask \leftarrow [Mask;\ TMask]$\;
		}\EndFor
		\State $P \leftarrow$ \textup{Generate a population with decision vectors taken from $Dec\cdot Mask$}\;
		\Return{\(P\)}
	\end{algorithm}
	% ========================== 算法4 ================================
        % ========================== 算法4 ===============================
	% ========================= 算法5 ======================================
	\indent To find promising variables, this paper proposes a function shown in Algorithm 5. As shown in Fig. 4, to avoid losing promising variables, the Pareto optimal solutions are selected from population $T$ to compute $initMask$. For each selected solution, we iteratively generate new solutions by setting the highest-scoring nonzero element in the mask vector to zero at each step. If the new solution dominates the original solutions, it indicates that the variables previously set to 0 are less important, and the operation is repeated. If an original solution dominates the new solution, it indicates that the variable set to 0 is important, and the loop operation is terminated. After repeating the above procedure for all Pareto optimal solutions in population $T$, a set of non-dominated solutions is obtained. Then sum their binary vectors element-wise and divide by the total number of non-dominated solutions to obtain \(initMask\). Each element of \(initMask\) represents the probability that the corresponding position element of the initial binary vector is 1.\\
    \\
    \begin{algorithm}[H]
		\caption{\(getMask(T, Score)\)}\label{alg:alg1}
		\SetKwData{Left}{left}\SetKwData{This}
		{this}\SetKwData{Up}{up}
		\SetKwFunction{Union}{Union}
		\SetKwFunction{FindCompress}
		{FindCompress} \SetKwInOut{Input}{Input}
		\SetKwInOut{Output}{Output}
		\Input{\(T\) (a population used to calculate the initial mask vectors), \(Score\) (scores of decision variables)}
		\Output{\(initMask\) (mask template for guiding initialization of initial mask vectors)}
		\State \(D \leftarrow \textup{Number of decision variables}\)\;
		\State \(initMask \leftarrow 1 \times D \textup{ vector of zeros}\)\;
		\State \textup{\(PS \leftarrow\) The Pareto optimal solutions of the population $T$}\;
		
		\For{\textup{\textbf{p}} \in \ $PS$}{
			
			\State \(Tmask\) \leftarrow \(\textup{\textbf{p}}.mask\)\;
			
			\While{$any\ element\ of\ \textup{\textbf{p}}.mask\ is\ nonzero$}{
                \State \textup{\(\textbf{p}' \leftarrow \textbf{p}\)}\;
				\State \textup{\(idx \leftarrow\) Select the index with the highest $Score$ among the nonzero elements of $\textbf{p}.mask$}\;
				
                \State \textup{\(\textbf{p}.mask_{idx} \leftarrow 0\)}\;
                
				\If{\(\textup{\textbf{p }} is\ dominated\ by\ \textup{\textbf{p}}'\)}{
					\tcp*[h]{Add mask and initMask element-wise}\\
					\State \(initMask\) \leftarrow \(initMask\) \textup{+} \(\textup{\textbf{p}}'.mask\)\;
					\textbf{\textup{break;}}
				}\Endif    
			}
		}\EndFor
		\tcp*[h]{Perform element-wise division of \(initMask\) by the size of \(PS\)}\\
		\State \(initMask \leftarrow initMask\ /\ |PS|\)\;
		
		\Return{\(initMask\)}
	\end{algorithm}
	\begin{figure}[]
		\centering
		% width=\textwidth 保证占满两栏
		\includegraphics[width=\textwidth]{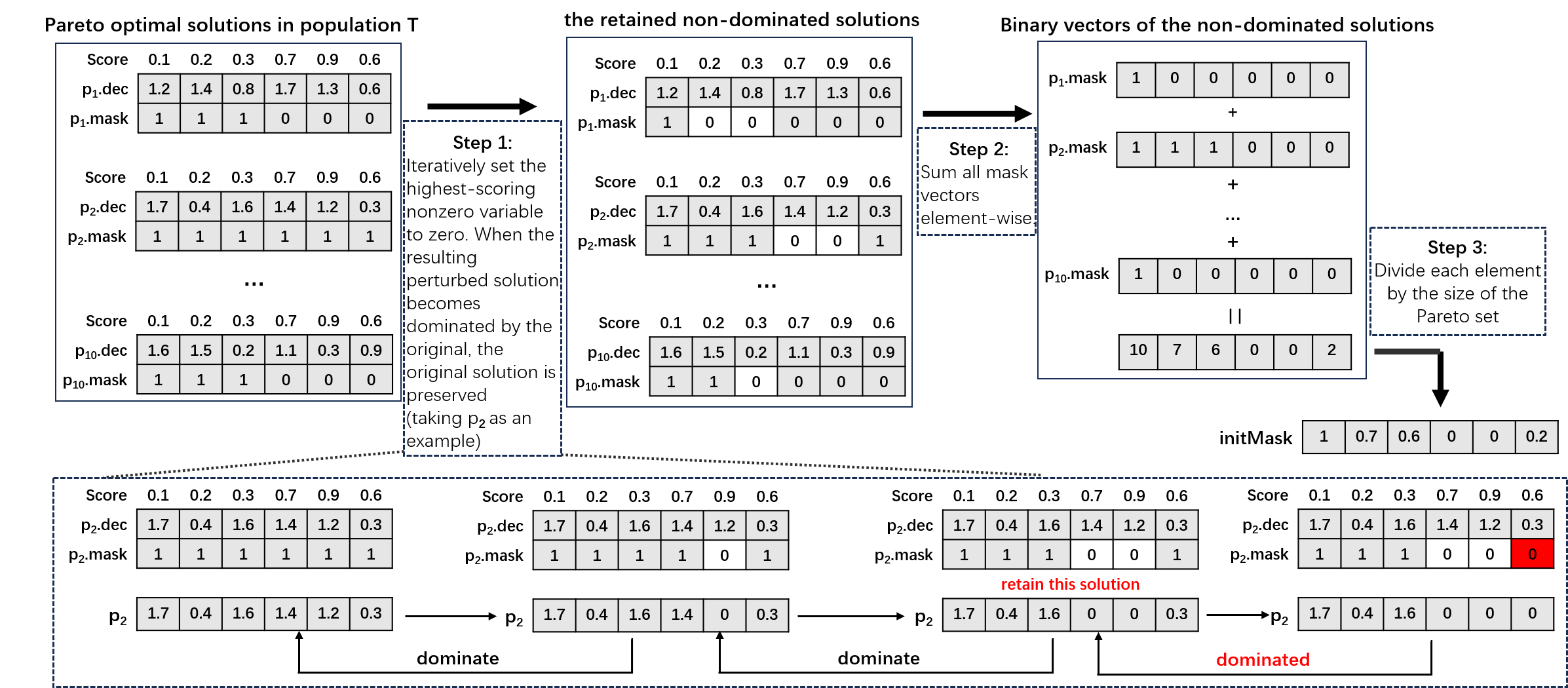}
		\caption{Example description of getting \(initMask\). Assume the Pareto set of population $T$ contains 10 solutions. In the retained non-dominated solutions, white entries in the mask vector $p_i.mask$ indicate variables that can be set to $0$ without causing the perturbed solution to become dominated by the original — i.e., they are unimportant.
        }
		\label{fig:wide-svg}
	\end{figure}
	\subsection{Genetic operators of SMOEA-OPS}
    \begin{algorithm}[H]
		\caption{\(Variation(P, Score,\lambda, flag)\)}\label{alg:alg1}
		\SetKwData{Left}{left}\SetKwData{This}
		{this}\SetKwData{Up}{up}
		\SetKwFunction{Union}{Union}
		\SetKwFunction{FindCompress}
		{FindCompress} \SetKwInOut{Input}{Input}
		\SetKwInOut{Output}{Output}
		\Input{\(P\) (current population), \(Score\) (scores of decision variables), \(\lambda\) (ratio of consumed evaluations), \(flag\) (flag indicating whether the \(initMask\) is valid) }
		\Output{ \(O\) (offspring population)}
		\State \(O \leftarrow \emptyset\)\;
		\While{P is not empty}{
			\State [\textbf{\textup{p, q}}] \leftarrow \textup{Randomly select two parents from \(P\)}\;
			\State \(P\) \leftarrow \(P\ \) \textbackslash \(\ \) [{\textbf{\textup{p, q}}}]\;
			\State \textup{\textbf{o}}\(.mask\) \leftarrow \(\) \textbf{\textup{p}}.\(mask\)\;
			\tcp*[h]{Crossover}\\
			\If{$rand()$ \leq \(0.5\)}{ 
				
                \State \textup{\(m \leftarrow\) Select one variable from the nonzero elements of \(\textbf{p}.mask \cap \textbf{q}.\overline{mask}\) using the selection probabilities defined by Eq.(9)}\;
                
				\State \textup{Set the \(m\)-th element in \textbf{o}.\(mask\) to 0}\;
			} \Else {
				\State \textup{\(m \leftarrow\) Select one variable from the nonzero elements of \(\textbf{p}.\overline{mask} \cap \textbf{q}.mask\) using the selection probabilities defined by Eq.(10)}\;
                
				\State \textup{Set the \(m\)-th element in \textbf{o}.\(mask\) to 1}\;
			} \Endif
			
			\tcp*[h]{Mutation}\\
			\If{$rand() \ge \lambda\ ||\ flag\ ==\ 1$}{
				\If{$rand()$ \leq \(0.5\)}{
					\State \textup{\(n \leftarrow\ \)Select one element from the nonzero elements in $\textbf{o}.mask$ according to the probability defined by Eq.(11)}\;
                    
					\State \textup{Set the \(n\)-th element in \textbf{o}\(.mask\) to 0}\;
				} \Else {
					\State \textup{\(n \leftarrow\ \)Select one element from the zero elements in $\textbf{o}.mask$ according to the probability defined by Eq.(12)}\;
					
					\State \textup{Set the \(n\)-th element in \textbf{o}\(.mask\) to 1}\;
				} \Endif
			}\Endif
			
			\State \textup{Update the mutation probability vector $mutProb$ using Eq.(14)}\;
			\tcp*[h]{Generate the real vector of offspring \textbf{o}}\\
            
            \State \textup{\(\textbf{o}.dec \leftarrow\) Perform simulated binary crossover based on \textbf{p}\(.dec\) and \textbf{q}\(.dec,\) and apply polynomial mutation with probabilities \(mutProb\)}\;
            
			\State \(O \leftarrow O\ \cup\ \{\textbf{o}\};\)
		}
		\Return{\(O\)}
	\end{algorithm}
	As shown in Algorithm 6, the genetic operators of SMOEA-OPS have a similar framework to SparseEA. First, randomly select two parents \textup{\textbf{p}} and \textup{\textbf{q}} from \(P\) to generate offspring \textup{\textbf{o}} each time, and set the binary vector of \textup{\textbf{o}} to be the same as that of \textup{\textbf{p}}. Then, one of the following two crossover operations is chosen with equal probability: select a nonzero element from $\textbf{p}.mask\cap \textbf{q}.\overline{mask}$, where the selection probability of each element is computed by Eq.(9). Then, set the corresponding element in $\textbf{o}.mask$ to $0$; or select a nonzero element from $\textbf{p}.\overline{mask}\cap \textbf{q}.mask$, where the selection probability of each element is computed by Eq.(10). Then, set the corresponding element in $\textbf{o}.mask$ to $1$. Specifically, during the crossover operation, the probability that the \(j\)-th element of offspring \textbf{o} is selected and assigned 0 can be calculated as follows:
	
	\begin{equation}
		P_{j,0}^{C} = \frac{Score_j}{\sum_{idx\in C_1} Score_{idx}}
		\label{eq:pc0}
	\end{equation}
	where \(C_1\) denotes the intersection of the nonzero elements between \textup{\textbf{p}}.\(mask\) and \textup{\textbf{q}}.\(\overline{mask}\). Meanwhile, the probability that the \(j\)-th element of offspring \textbf{o} is selected and assigned 1 can be calculated as follows:
	\begin{equation}
		P_{j,1}^{C} = \frac{\frac{1}{Score_j}}{\sum_{idx\in C_0} \frac{1}{Score_{idx}}}
		\label{eq:pc0}
	\end{equation}
	where \(C_0\) denotes the intersection of the nonzero elements between \textup{\textbf{p}}.\(\overline{mask} \) and \textup{\textbf{q}}.\(mask\).\\ 
    \indent Since the mask vectors obtained after population initialization can accurately activate most of the important variables, the mutation probability of mask vectors can be gradually reduced as evolution progresses, which also helps optimize the real variables and reduces computational cost. Specifically, the mutation probability of mask vectors is set to dynamically change as $1 - \lambda$, where $\lambda$ is the ratio of consumed evaluations, i.e., $FE/FE_{max}$.
    % ============================== 算法5 ============================
    \indent When \textbf{o}.mask mutates, one of the following two operations is performed with the same probability: select an element from the nonzero entries of \textup{\textbf{o}}.\(mask\) with probabilities given by Eq.(11), and set it to 0; or select an element from the nonzero entries of \(\textup{\textbf{o}}.\overline{mask}\) with probabilities given by Eq.(12), and set it to 1. Specifically, during the mutation operation, the probability that the \(j\)-th element of offspring \textbf{o} is selected and assigned 0 can be calculated as follows:
	\begin{equation}
		P_{j,0}^{M} = \frac{Score_j}{\sum_{idx\in M_1} Score_{idx}}
		\label{eq:pc0}
	\end{equation}
	where \(M_1\) denotes the set of nonzero elements from \textbf{o}.\(mask\). Meanwhile, the probability that the \(j\)-th element is selected and assigned 1 can be calculated as follows:
	\begin{equation}
		P_{j,1}^{M} = \frac{\frac{1}{Score_j}}{\sum_{idx\in M_0} \frac{1}{Score_{idx}}}
		\label{eq:pc0}
	\end{equation}
	where \(M_0\) denotes the set of zero elements from \textbf{o}.\(mask\).\\
	\indent The real vector \textbf{o}\(.dec\) of solution \textbf{o} is generated by simulated binary crossover~\cite{deb1995simulated} and polynomial mutation~\cite{deb1996combined}, similar to most MOEA. In this paper, the mutation probability of each real variable is proportional to the standard deviation of values in the corresponding column of \(IntervalFronts\). The calculation of $IntervalFronts$ is described in Lines 2–16 of Algorithm 3. The formula for calculating the standard deviation of the elements in the \(j\)-th column of \(IntervalFronts\) is as follows:
	{\fontsize{7.5pt}{9pt}\selectfont
    \begin{equation}
	std_j = \sqrt{\frac{1}{NVal-1}\sum_{i=1}^{NVal}{(IntervalFronts_{i,j} - \overline{IntervalFronts_{:,j}})^2}}
		\label{eq:std}
    \end{equation}
    }\\
    \noindent where \(\overline{IntervalFronts_{:,j}}\) represents the average value of all the elements in the \(j\)-th column of the matrix \(IntervalFronts\). Therefore, the formula for calculating the mutation probability of the \(j\)-th real variable in the solution \textbf{r} can be written as follows:
	
	\begin{equation}
		mutProb_{\textbf{r},j} = 
		\begin{cases}
			\frac{std_j}{\sum_{k\in nz\_idx}std_k},  &  \textbf{r}.mask_{j} \neq 0,\\[6pt]
			0 & \textbf{r}.mask_{j} = 0
		\end{cases}
		\label{eq:mut}
	\end{equation}
    where \(nz\_idx\) denotes the set of indices of nonzero elements from \textbf{r}\(.mask\). As the population evolves, the number of nonzero elements gradually decreases. The term $\sum_{k\in nz\_idx} std_k$, which excludes the standard deviations of the zero elements in the current solution, increases the mutation probability of all nonzero variables in solution \textbf{r}. If the $j$-th element of the mask vector of \textbf{r} is $0$, the corresponding real variable does not need to mutate, and its mutation probability is therefore set to $0$.\\
    \indent This probability scheme also prevents real variables from being sampled only within the intervals with the smallest non-dominated front numbers, thereby improving diversity and reducing the risk of getting trapped in local optima.
	% ===========================================================
	% ===========================================================
	\subsection{The proposed Pareto-guided resampling strategy for optimizing real vectors}
	Many LSSMOOAs focus only on the optimization of binary vectors and ignore the optimization of real vectors, resulting in poor performance. Therefore, this paper proposes a Pareto-guided resampling method based on the normal distribution for optimizing real vectors, as shown in Algorithm 7. When dealing with LSMOPs, their complex search landscapes typically require strong early-stage exploration to maintain diversity. Therefore, the evolutionary process is divided into two stages according to the number of decision variables. In the first stage, the population's exploration ability is maintained, whereas in the second stage, its convergence toward the Pareto set is further strengthened.
    % ============================= 算法7 ==============================
	\begin{algorithm}[H]
		\caption{\(PGResample(P, \mu, \sigma, \lambda)\)}\label{alg:alg1}
		\SetKwData{Left}{left}\SetKwData{This}
		{this}\SetKwData{Up}{up}
		\SetKwFunction{Union}{Union}
		\SetKwFunction{FindCompress}
		{FindCompress} \SetKwInOut{Input}{Input}
		\SetKwInOut{Output}{Output}
		\Input{\(P\) (offspring population), \(\mu\) (means of variables in Pareto optimal solutions), \(\sigma\) (standard deviations of variables in Pareto optimal solutions), \(\lambda\) (ratio of consumed evaluations)}
		\Output{\(Q\) (resampled population)}
		\State \(Q \leftarrow \emptyset\)\;
		\State \(D \leftarrow \textup{the number of decision variables}\)\;
		\For{\(\textup{\textbf{p}} \in P\)}{
			\If{$\lambda$ $\leq$ $1-\frac{lnD}{\sqrt{D}}$}{
				\State \textup{\(m \leftarrow\) Select a variable from \textbf{p} with probability calculated by Eq.(15)}\;
                $\textbf{p}.dec_m \leftarrow$ Generate a random sample from $\mathcal{N}(\mu_m,\ \sigma_m^2)$\;
			}\Else{
                \For{$j = 1\ to\ D$} {
                     \If{$\textup{\textbf{p}}.dec_j\textup{ is outside }[\mu_j-2\sigma_j,\mu_j+2\sigma_j]$}{
                        \textup{\textbf{p}}$.dec_j \leftarrow$ Generate a random sample from $\mathcal{N}(\mu_j,\ \sigma_j^2)$\;
                    }   
                }      
			}\Endif
			\State \(Q \leftarrow Q\ \cup\ \{\textbf{p}\}\)\;
		}
		\Return{\(Q\)}
	\end{algorithm}
	% ===========================================================
    
	In the early stage, for each solution, a variable is selected with the probability proportional to its standard deviation in the current population. The probability of the \(i\)-th variable being selected is calculated as follows:
	\begin{equation}
		resampleProb_i = \frac{\sigma_i}{\sum_{j=1}^{D}\sigma_j}
		\label{eq:std}
	\end{equation}
	where $\sigma_i$ and \(D\) represent the standard deviation of the \(i\)-th variables in Pareto optimal solutions and the number of decision variables, respectively. In the first stage, if the \(i\)-th variable is selected in a solution, draw a sample from the normal distribution \(\mathcal{N}(\mu_i,\sigma_i^2)\) and assign it to the \(i\)-th variable. Therefore, in the first stage, by resampling only those offspring's real variables that deviate excessively from the Pareto set, the population's diversity is preserved. In the second stage, all real variables of the population learn from the current Pareto set (Algorithm~7, lines 8–10): all real variables that deviate from the Pareto set are resampled to strengthen offspring convergence toward the Pareto set. Specifically, for each solution, if there exists \(j\in\{1,\dots,D\}\) such that the \(j\)-th variable lies outside the interval \([\mu_j-2\sigma_j,\ \mu_j+2\sigma_j]\), then resample the \(j\)-th variable from \(\mathcal{N}(\mu_j,\sigma_j^2)\).\\
	\indent As the number of decision variables grows, locating Pareto-optimal solutions becomes more difficult. To increase the chance of escaping local optima, we therefore extend the exploration phase for higher-dimensional problems. Concretely, we split the evolutionary process so that the initial $(1 - \frac{lnD}{\sqrt{D}})\%$ of the run focuses on the exploration, while the final $\frac{lnD}{\sqrt{D}}\%$ of the run concentrates on the exploitation. We reserve the final $\frac{lnD}{\sqrt{D}}\%$ of the evolutionary process for exploitation because this ratio can adaptively change as the dimension changes: $lnD$ grows slowly, while $\sqrt{D}$ grows faster than $lnD$ but not as fast as \(D\), causing the overall ratio to shrink with \(D\) without collapsing too abruptly. Thus, in higher-dimensional problems, the exploration stage can be extended to improve the chances of finding high-quality solutions. Moreover, as \(D\) varies across problems, $\frac{lnD}{\sqrt{D}}$ remains within a reasonable range.
\section{Experimental results}
    To evaluate the performance of SMOEA-OPS, eight benchmark problems, namely SMOP1-SMOP8~\cite{tian2020evolutionary}, and three real-world SMOPs are used to verify the obtained solutions. Real-world SMOPs introduce problems with real-valued decision spaces and binary decision spaces, namely, sparse signal reconstruction (SR)~\cite{li2018preference}, community detection (CD)~\cite{ye2022solving}, and the knapsack problem (KP)~\cite{su2022comparing}. Moreover, SparseEA2~\cite{SparseEA2}, S-NSGA-II~\cite{kropp2023improved}, MSKEA~\cite{ye2024mskea}, TELSO~\cite{qi2024two}, MGCEA~\cite{MGCEA} are selected as baseline algorithms. All experiments were conducted on PlatEMO~\cite{tian2023practical}.\\
	\subsection{Settings of algorithms}
	The parameters of the compared MOEAs are set to the values recommended in their original papers. For SparseEA2, the number of groups is set to 4. For MGCEA and the proposed SMOEA-OPS, the number of intervals \(NV al\) is set to 5, and the number of samplings per interval \(SV al\) is set to 2.\\
    \indent Simulated binary crossover~\cite{deb1995simulated} and polynomial mutation~\cite{deb1996combined} are used to generate the real variables of offspring solutions for SparseEA2, MSKEA, MGCEA, and the proposed SMOEA-OPS. For algorithms other than SMOEA-OPS, the crossover and mutation probabilities are set to \(\,1\,\) and \(\,1/D\,\), respectively. The distribution indices for crossover and mutation are both set to 20. S-NSGA-II and TELSO use their own genetic operators to generate real variables for offspring solutions. In particular, S-NSGA-II generates real variables within [0, 1] and rounds them to obtain binary variables~\cite{MGCEA}. \\
	\indent The population size is set to 100 in both the benchmark SMOPs and the real-world SMOPs. The maximum number of function evaluations is set to $100 \times D$ on benchmark problems. For real-world SMOPs, the maximum number of function evaluations is set to \(20\times D\) for SR1-SR4, 20000 for CD1-CD4, and \(100\times D\) for KP1-KP4.
	\subsection{Settings of problems}
    \setlength{\tabcolsep}{4pt} % 根据需要微调列间距
	\begin{table*}[H]
		\centering
		\scriptsize
		\renewcommand{\arraystretch}{1.2}
		\caption{Parameter settings of eight benchmark SMOPs and three real-world SMOPs.}
		\label{tab:smop-and-sr-benchmarks}
		\begin{tabular*}{\textwidth}{@{\extracolsep{\fill}} c c c c c c c @{} }
			\toprule
			\makecell{Benchmark\\problem}
			& \makecell{Type of\\variables}
			& \makecell{No. of\\variables}
			& \makecell{No. of\\objectives}
			& \makecell{Sparsity of\\Pareto optimal solutions}
			& {} % 占位列
			& {} % 占位列, 保持7列结构
			\\
			\midrule
			% SMOP 部分
			\multirow{4}{*}{SMOP1--SMOP8}
			& \multirow{4}{*}{Real}
			& 100   & \multirow{4}{*}{2}   & \multirow{4}{*}{0.1} & & \\
			&       & 500   &                   &                     & & \\
			&       & 1000  &                   &                     & & \\
			&       & 5000  &                   &                     & & \\
			\midrule
			% SR 部分
			\makecell{Sparse signal\\reconstruction problem}
			& \makecell{Type of\\variables}
			& \makecell{No. of\\variables}
			& Dataset
			& \makecell{Length of\\signal}
			& \makecell{Length of received\\signal}
			& \makecell{Sparsity of\\signal} \\
			\midrule
			SR1 & \multirow{4}{*}{Real}
			& 1024   & Synthetic~\cite{li2018preference} & 1024   & 512   & 256   \\
			SR2 &                     & 2048  & Synthetic~\cite{li2018preference} & 2048  & 1024   & 512   \\
			SR3 &                     & 3072  & Synthetic~\cite{li2018preference} & 3072  & 1536  & 768  \\
			SR4 &                     & 4096 & Synthetic~\cite{li2018preference} & 4096 & 2048  & 1024  \\
			\midrule
			% CD 部分
			\makecell{Community detection\\problem}
			& \makecell{Type of\\variables}
			& \makecell{No. of\\variables}
			& Dataset
			& \makecell{No. of nodes}
			& \makecell{No. of edges} \\
			\midrule
			CD1 & \multirow{4}{*}{Binary}
			& 34    & Karate\textsuperscript{1}      & 34    & 78    \\
			CD2 &                     & 62    & Dolphin\textsuperscript{1}     & 62    & 159   \\
			CD3 &                     & 105   & Polbook\textsuperscript{1}     & 105   & 441   \\
			CD4 &                     & 115   & Football\textsuperscript{1}    & 115   & 613   \\
			\midrule
			% KP 部分
			\makecell{Knapsack problem}
			& \makecell{Type of\\variables}
			& \makecell{No. of\\variables}
			& Dataset
			& \makecell{No. of items}
			& {} % 占位列
			& {} % 占位列
			\\
			\midrule
			KP1 & \multirow{4}{*}{Binary}
			& 100   & Synthetic~\cite{su2022comparing}    & 100   & & \\
			KP2 &                     & 500   & Synthetic~\cite{su2022comparing}    & 500   & & \\
			KP3 &                     & 1000  & Synthetic~\cite{su2022comparing}    & 1000  & & \\
			KP4 &                     & 5000  & Synthetic~\cite{su2022comparing}    & 5000  & & \\
			\bottomrule
		\end{tabular*}
        % =====================================
        \vspace{0.2cm} % 表格和注释之间加点间距
  \begin{minipage}{0.9\linewidth} % 控制宽度
    \raggedright
    \footnotesize
    \textsuperscript{1} http://www-personal.umich.edu/$\sim$mejn/netdata/ \quad
  \end{minipage}
        % =====================================
	\end{table*}
    \setlength{\tabcolsep}{1pt}
	\begin{table*}[]
		\centering
		\caption{IGD values obtained by SparseEA2, MSKEA, S-NSGA-II, TELSO, MGCEA and the proposed SMOEA-OPS on SMOP1–SMOP8 with 100 to 5000 decision variables and 2 objectives.
		}
		\begin{minipage}{\textwidth}
			\scriptsize
			\renewcommand{\arraystretch}{1.3}
			\begin{tabular}{M{1.1cm}M{0.7cm}*{5}{M{2.5cm}}M{2.2cm}}
				\toprule
				\multicolumn{1}{c}{\textbf{Problem}} 
				& \multicolumn{1}{c}{\textbf{D}} 
				& \multicolumn{1}{c}{\textbf{SparseEA2}} 
				& \multicolumn{1}{c}{\textbf{MSKEA}} 
				& \multicolumn{1}{c}{\textbf{S-NSGA-II}} 
				& \multicolumn{1}{c}{\textbf{TELSO}} 
				& \multicolumn{1}{c}{\textbf{MGCEA}} 
				& \multicolumn{1}{c}{\textbf{SMOEA-OPS}} \\
				\midrule
				\multirow{4}{*}{\centering SMOP1}
				& 100 
				& 5.7122e-3 (5.53e-4) $-$ & 4.5016e-3 (1.74e-3) $-$ & 4.9017e-3 (6.34e-4) $-$ & 7.8439e-2 (1.11e-4) $-$ & 3.7469e-3 (4.82e-5) $-$ & \cellcolor{lightgray}3.6989e-3 (1.55e-5) \\
				& 500 
				& 6.5295e-3 (5.23e-4) $-$ & 8.9548e-3 (2.97e-3) $-$ & 9.0297e-3 (3.51e-3) $-$ & 7.7230e-2 (2.05e-3) $-$ & 3.7854e-3 (1.83e-4) $-$ & \cellcolor{lightgray}3.7074e-3 (2.38e-5) \\
				& 1000 
				& 7.1441e-3 (6.03e-4) $-$ & 1.3875e-2 (3.26e-3) $-$ & 1.2192e-2 (3.93e-3) $-$ & 7.5729e-2 (4.04e-3) $-$ & 4.8103e-3 (1.63e-3) $-$ & \cellcolor{lightgray}3.7081e-3 (2.38e-5) \\  
				& 5000 
				& 3.2256e-2 (1.06e-3) $-$ & 2.7706e-2 (1.22e-3) $-$ & 2.0920e-2 (4.97e-3) $-$ & 7.4069e-2 (5.50e-3) $-$ & 7.5553e-3 (4.45e-3) $-$ & \cellcolor{lightgray}3.7076e-3 (2.37e-5) \\
				\midrule
				\multirow{4}{*}{\centering SMOP2}
				& 100 
				& 1.1063e-2 (4.73e-3) $-$ & 8.0084e-3 (4.36e-3) $-$ & 1.4832e-2 (9.03e-3) $-$ & 1.6283e-1 (1.80e-6) $-$ & 4.1090e-3 (7.37e-4) $-$ & \cellcolor{lightgray}3.7031e-3 (1.77e-5) \\
				& 500 
				& 1.2190e-2 (3.88e-3) $-$ & 2.8170e-2 (7.59e-3) $-$ & 2.7790e-2 (1.05e-2) $-$ & 1.6027e-1 (5.03e-3) $-$ & 3.8687e-3 (2.75e-4) $-$ & \cellcolor{lightgray}3.7486e-3 (6.81e-5) \\
				& 1000 
				& 1.4147e-2 (2.66e-3) $-$ & 4.2129e-2 (5.92e-3) $-$ & 4.2559e-2 (1.27e-2) $-$ & 1.5953e-1 (6.17e-3) $-$ & 4.8629e-3 (1.28e-3) $-$ & \cellcolor{lightgray}3.9351e-3 (1.71e-4) \\
				& 5000 
				& 7.2252e-2 (3.15e-3) $-$ & 7.2474e-2 (2.98e-3) $-$ & 8.6190e-2 (1.43e-2) $-$ & 1.6051e-1 (3.66e-3) $-$ & 1.7598e-2 (2.39e-3) $-$ & \cellcolor{lightgray}6.1300e-3 (4.19e-4) \\
				\midrule
				\multirow{4}{*}{\centering SMOP3}
				& 100 
				& 7.1524e-3 (1.62e-3) $-$ & 3.9728e-3 (9.68e-4) $-$ & 9.1014e-3 (5.56e-3) $-$ & 7.8268e-2 (1.02e-3) $-$ & 3.7096e-3 (2.34e-5) = & \cellcolor{lightgray}3.7002e-3 (1.68e-5) \\
				& 500 
				& 8.4182e-3 (1.70e-3) $-$ & 6.5042e-3 (1.90e-3) $-$ & 1.1869e-2 (3.43e-3) $-$ & 7.7729e-2 (3.26e-4) $-$ & 3.7018e-3 (1.69e-5) = & \cellcolor{lightgray}3.7014e-3 (1.91e-5) \\
				& 1000 
				& 1.0899e-2 (1.36e-3) $-$ & 1.0058e-2 (2.19e-3) $-$ & 1.8050e-2 (3.62e-3) $-$ & 7.7192e-2 (2.95e-3) $-$ &\cellcolor{lightgray} 3.7034e-3 (2.00e-5) = & 3.7148e-3 (5.59e-5) \\
				& 5000 
				& 3.2756e-2 (1.30e-3) $-$ & 2.2859e-2 (1.13e-3) $-$ & 2.3058e-2 (3.50e-3) $-$ & 7.6699e-2 (3.79e-3) $-$ &\cellcolor{lightgray} 3.7272e-3 (7.49e-5) = & 4.0860e-3 (1.11e-3) \\
				\midrule
				\multirow{4}{*}{\centering SMOP4}
				& 100 
				& 4.6913e-3 (1.74e-4) + & 4.1131e-3 (8.46e-5) + & 5.3192e-3 (4.01e-4) $-$ &\cellcolor{lightgray} 3.9288e-3 (6.53e-6) + & 4.1368e-3 (6.90e-5) = & 4.8039e-3 (1.21e-3) \\
				& 500 
				& 4.7673e-3 (2.98e-4) $-$ & 4.1329e-3 (6.74e-5) = & 4.8457e-3 (2.11e-4) $-$ &\cellcolor{lightgray} 3.9275e-3 (2.37e-6) + & 4.0980e-3 (6.99e-5) = & 4.1020e-3 (6.83e-5) \\
				& 1000 
				& 4.8051e-3 (2.87e-4) $-$ & 4.1503e-3 (6.44e-5) = & 4.8746e-3 (2.89e-4) $-$ &\cellcolor{lightgray} 3.9287e-3 (1.96e-6) + & 4.1108e-3 (5.58e-5) = & 4.1183e-3 (7.30e-5) \\
				& 5000 
				& 4.7909e-3 (3.08e-4) $-$ & 4.1228e-3 (4.84e-5) = & 4.8435e-3 (2.05e-4) $-$ &\cellcolor{lightgray} 3.9297e-3 (6.74e-7) + & 4.1144e-3 (6.66e-5) = & 4.1307e-3 (6.89e-5) \\
				\midrule
				\multirow{4}{*}{\centering SMOP5}
				& 100 
				& 5.5681e-3 (2.66e-4) $-$ & 4.4520e-3 (3.13e-4) $-$ & 4.9248e-3 (1.99e-4) $-$ & 8.7845e-3 (4.93e-3) $-$ & 4.9276e-3 (3.51e-4) $-$ & \cellcolor{lightgray}4.2897e-3 (2.17e-4) \\
				& 500 
				& 5.3210e-3 (2.13e-4) $-$ & 4.2066e-3 (1.36e-4) = & 4.8926e-3 (2.31e-4) $-$ & 8.4651e-3 (4.58e-3) $-$ & 4.6166e-3 (1.47e-4) $-$ & \cellcolor{lightgray}4.1515e-3 (6.49e-5) \\
				& 1000 
				& 5.2531e-3 (1.87e-4) $-$ & 4.1869e-3 (6.98e-5) $-$ & 4.9299e-3 (3.56e-4) $-$ & 1.0495e-2 (7.96e-3) $-$ & 4.2561e-3 (8.46e-5) $-$ & \cellcolor{lightgray}4.1431e-3 (7.38e-5) \\
				& 5000 
				& 5.0780e-3 (2.04e-4) $-$ & 4.2726e-3 (8.07e-5) $-$ & 2.1247e-2 (3.23e-3) $-$ & 1.1361e-2 (9.53e-3) $-$ &\cellcolor{lightgray} 4.1154e-3 (6.61e-5) = & 4.1441e-3 (6.95e-5) \\
				\midrule
				\multirow{4}{*}{\centering SMOP6}
				& 100 
				& 6.8374e-3 (4.22e-4) $-$ & 4.7155e-3 (5.69e-4) = & 4.9331e-3 (1.67e-4) $-$ & 3.2316e-2 (9.81e-3) $-$ & 4.7717e-3 (2.52e-4) $-$ & \cellcolor{lightgray}4.5788e-3 (2.12e-4) \\
				& 500 
				& 6.5316e-3 (3.39e-4) $-$ & 4.4555e-3 (1.69e-4) $-$ & 4.7642e-3 (1.63e-4) $-$ & 2.6963e-2 (1.41e-2) $-$ & 4.3022e-3 (9.11e-5) $-$ & \cellcolor{lightgray}4.1992e-3 (7.40e-5) \\
				& 1000 
				& 6.6725e-3 (3.44e-4) $-$ & 4.5332e-3 (1.33e-4) $-$ & 4.8972e-3 (3.03e-4) $-$ & 2.8363e-2 (1.42e-2) $-$ & 4.2148e-3 (5.83e-5) = & \cellcolor{lightgray}4.2118e-3 (7.37e-5) \\
				& 5000 
				& 6.9600e-3 (3.25e-4) $-$ & 4.9903e-3 (9.38e-5) $-$ & 8.0920e-3 (2.65e-3) $-$ & 2.5303e-2 (1.57e-2) $-$ &\cellcolor{lightgray} 4.2234e-3 (7.45e-5) + & 4.3515e-3 (5.49e-5) \\
				\midrule
				\multirow{4}{*}{\centering SMOP7}
				& 100 
				& 1.5528e-2 (5.94e-3) $-$ & 1.0214e-2 (8.00e-3) $-$ & 1.9755e-2 (1.38e-2) $-$ & 2.2918e-1 (1.72e-3) $-$ & 6.9525e-3 (3.02e-3) $-$ & \cellcolor{lightgray}4.1308e-3 (4.85e-5) \\
				& 500 
				& 8.1177e-3 (2.09e-3) $-$ & 3.8832e-2 (7.44e-3) $-$ & 4.6095e-2 (1.97e-2) $-$ & 2.2570e-1 (7.10e-3) $-$ & 1.3803e-2 (6.26e-3) $-$ & \cellcolor{lightgray}4.1853e-3 (1.65e-4) \\
				& 1000 
				& 1.0003e-2 (2.03e-3) $-$ & 5.5188e-2 (8.28e-3) $-$ & 7.4755e-2 (1.98e-2) $-$ & 2.2680e-1 (4.89e-3) $-$ & 2.8040e-2 (8.02e-3) $-$ & \cellcolor{lightgray}4.4035e-3 (2.74e-4) \\
				& 5000 
				& 6.2637e-2 (2.88e-3) $-$ & 9.1474e-2 (3.01e-3) $-$ & 1.2144e-1 (1.99e-2) $-$ & 2.1850e-1 (1.80e-2) $-$ & 5.2130e-2 (2.14e-2) $-$ & \cellcolor{lightgray}6.6589e-3 (6.61e-4) \\
				\midrule
				\multirow{4}{*}{\centering SMOP8}
				& 100 
				& 1.3612e-1 (2.42e-2) $-$ & 1.0684e-1 (2.68e-2) $-$ & 1.1844e-1 (3.32e-2) $-$ & 5.1247e-1 (1.84e-3) $-$ & 9.0671e-2 (2.09e-2) $-$ & \cellcolor{lightgray}7.4686e-2 (2.27e-2) \\
				& 500 
				& 1.8353e-1 (1.05e-2) $-$ & 1.5225e-1 (1.81e-2) $-$ & 1.5038e-1 (1.90e-2) $-$ & 5.1324e-1 (5.50e-3) $-$ & 1.1410e-1 (9.16e-3) $-$ & \cellcolor{lightgray}1.0122e-1 (1.51e-2) \\
				& 1000 
				& 1.9975e-1 (1.23e-2) $-$ & 1.8052e-1 (1.38e-2) $-$ & 1.6985e-1 (2.13e-2) $-$ & 5.1192e-1 (1.23e-2) $-$ &\cellcolor{lightgray} 1.1858e-1 (8.39e-3) + & 1.4100e-1 (2.09e-2) \\
				& 5000 
				& 2.5313e-1 (4.70e-3) $-$ & 2.5500e-1 (8.45e-3) $-$ & 2.1290e-1 (1.15e-2) + & 5.0885e-1 (1.54e-2) $-$ &\cellcolor{lightgray} 1.5993e-1 (7.56e-3) + & 2.3023e-1 (1.45e-2) \\
				\midrule
				\multicolumn{2}{c}{\textbf{$+/-/\approx$}} 
				& \multicolumn{1}{c}{1/31/0} 
				& \multicolumn{1}{c}{1/26/5} 
				& \multicolumn{1}{c}{1/31/0} 
				& \multicolumn{1}{c}{4/28/0} 
				& \multicolumn{1}{c}{3/19/10} 
				& \\
				\bottomrule
			\end{tabular}
			\label{tab:igd_comparison}
		\end{minipage}
	\end{table*}
	For the eight benchmark problems SMOP1–SMOP8, the number of objectives is set to 2, the number of decision variables \(D\) is set to 100, 500, 1000, and 5000, and the sparsity of Pareto optimal solutions is set to 0.1. The three real-world SMOPs are included in PlatEMO 4.8. Details can be found in Table 1.\\
    \indent To evaluate the performance of the algorithms, inverted generational distance (IGD)~\cite{bosman2003balance} and hypervolume (HV)~\cite{zitzler1999multiobjective} are used to assess the performance of each algorithm in 30 independent runs on benchmark problems and practical problems, respectively. As shown in Table 2, “$+$” indicates that an algorithm performs better than the proposed SMOEA-OPS, “$-$” indicates that an algorithm performs worse than the proposed SMOEA-OPS, and “$\approx$” indicates that an algorithm performs similarly to the proposed SMOEA-OPS. 
    \begin{table*}[]
		\centering
		\caption{HV values obtained by SparseEA2, MSKEA, S-NSGA-II, TELSO, MGCEA and the proposed SMOEA-OPS on SR, KP, and CD problems from 30 runs.}
		\begin{minipage}{\textwidth}
			\scriptsize
			\renewcommand{\arraystretch}{1.3}
			\begin{tabular}{M{1.1cm}M{0.7cm}*{5}{M{2.5cm}}M{2.2cm}}
				\toprule
				\multicolumn{1}{c}{\textbf{Problem}} 
				& \multicolumn{1}{c}{\textbf{D}} 
				& \multicolumn{1}{c}{\textbf{SparseEA2}} 
				& \multicolumn{1}{c}{\textbf{MSKEA}} 
				& \multicolumn{1}{c}{\textbf{S-NSGA-II}} 
				& \multicolumn{1}{c}{\textbf{TELSO}} 
				& \multicolumn{1}{c}{\textbf{MGCEA}} 
				& \multicolumn{1}{c}{\textbf{SMOEA-OPS}} \\
				\midrule
				\multirow{1}{*}{\centering SR1}
				& 1024 
				& 2.8183e-1 (6.08e-3) $-$ & 2.1159e-1 (9.03e-3) $-$ & 1.0328e-1 (3.66e-3) $-$ & 1.5643e-1 (8.57e-3) $-$ & 2.9988e-1 (5.18e-3) $-$ & \cellcolor{lightgray}3.2377e-1 (1.90e-3) \\
				
				\multirow{1}{*}{\centering SR2}
				& 2048 
				& 2.7606e-1 (5.54e-3) $-$ & 2.0086e-1 (6.57e-3) $-$ & 9.8417e-2 (2.54e-3) $-$ & 1.4440e-1 (7.64e-3) $-$ & 3.0254e-1 (3.81e-3) $-$ & \cellcolor{lightgray}3.2724e-1 (1.36e-3) \\
				
				\multirow{1}{*}{\centering SR3}
				& 3072 
				& 2.8286e-1 (4.79e-3) $-$ & 1.9495e-1 (3.53e-3) $-$ & 9.6735e-2 (2.01e-3) $-$ & 1.4764e-1 (6.68e-3) $-$ & 3.0979e-1 (3.55e-3) $-$ & \cellcolor{lightgray}3.4262e-1 (1.08e-3) \\
				
				\multirow{1}{*}{\centering SR4}
				& 4096 
				& 2.7242e-1 (3.85e-3) $-$ & 1.8447e-1 (5.14e-3) $-$ & 9.5077e-2 (1.69e-3) $-$ & 1.4123e-1 (5.70e-3) $-$ & 2.9758e-1 (2.59e-3) $-$ & \cellcolor{lightgray}3.2608e-1 (9.59e-4) \\
				
				\midrule
				\multirow{1}{*}{\centering KP1}
				& 100 
				& 7.8113e-2 (6.76e-3) $-$ & 8.9635e-2 (2.11e-3) $-$ & 7.8886e-2 (6.34e-3) $-$ & 4.2475e-2 (4.84e-3) $-$ & 7.6072e-2 (7.15e-3) $-$ & \cellcolor{lightgray}9.1270e-2 (1.24e-3) \\
				
				\multirow{1}{*}{\centering KP2}
				& 500 
				& 5.1618e-2 (4.41e-3) $-$ & 6.7841e-2 (1.43e-3) $-$ & 5.7868e-2 (3.24e-3) $-$ & 3.9174e-2 (3.38e-3) $-$ & 5.7051e-2 (4.67e-3) $-$ & \cellcolor{lightgray}6.8845e-2 (1.09e-3) \\
				
				\multirow{1}{*}{\centering KP3}
				& 1000 
				& 4.5920e-2 (5.26e-3) $-$ & 6.9031e-2 (1.35e-3) $-$ & 5.4462e-2 (2.86e-3) $-$ & 3.7677e-2 (4.38e-3) $-$ & 5.3129e-2 (5.74e-3) $-$ & \cellcolor{lightgray}7.2461e-2 (1.02e-3) \\
				
				\multirow{1}{*}{\centering KP4}
				& 5000 
				& 3.9439e-2 (4.62e-3) $-$ & 6.7897e-2 (6.51e-4) $-$ & 4.4923e-2 (1.20e-3) $-$ & 3.7900e-2 (2.45e-3) $-$ & 4.8013e-2 (2.86e-3) $-$ & \cellcolor{lightgray}6.8372e-2 (6.15e-4) \\
				
				\midrule
				\multirow{1}{*}{\centering CD1}
				& 34 
				& 7.9430e-1 (1.09e-2) $-$ & 8.1375e-1 (5.93e-4) $-$ & 7.2419e-1 (3.78e-2) $-$ & 6.8230e-1 (1.88e-2) $-$ & 7.9271e-1 (7.83e-3) $-$ & \cellcolor{lightgray}8.1452e-1 (2.52e-4) \\
				
				\multirow{1}{*}{\centering CD2}
				& 62 
				& 7.9568e-1 (8.96e-3) $-$ & 8.1983e-1 (1.36e-3) $-$ & 6.8152e-1 (3.37e-2) $-$ & 6.9517e-1 (1.09e-2) $-$ & 7.7649e-1 (1.10e-2) $-$ & \cellcolor{lightgray}8.2090e-1 (9.57e-4) \\
				
				\multirow{1}{*}{\centering CD3}
				& 105 
				& 7.6324e-1 (9.14e-3) $-$ & 7.9945e-1 (2.70e-3) $-$ & 4.9504e-1 (4.94e-2) $-$ & 6.4116e-1 (8.50e-3) $-$ & 7.3807e-1 (1.42e-2) $-$ & \cellcolor{lightgray}8.0157e-1 (2.20e-3) \\
				
				\multirow{1}{*}{\centering CD4}
				& 115 
				& 7.7945e-1 (3.79e-3) $-$ &\cellcolor{lightgray} 7.9110e-1 (7.46e-4) = & 6.2708e-1 (2.93e-2) $-$ & 7.2406e-1 (4.63e-3) $-$ & 7.5588e-1 (7.04e-3) $-$ & 7.9106e-1 (5.43e-4) \\
				
				\bottomrule
				\multirow{1}{*}{\centering + / - / $\approx$}
				& &0/12/0 &0/11/1 &0/12/0 &0/12/0 &0/12/0 & \\
				\midrule
			\end{tabular}
			\label{tab:igd_real}
		\end{minipage}
	\end{table*}
    \subsection{Comparative experiments}
	\begin{figure*}[]
		\centering
		% ========== 调整这里 ==========
		% gap: 列间固定间距（可设为 0pt, 0.5pt, 1pt, 0.005\textwidth 等）
		\newlength{\gap}
		\setlength{\gap}{-10pt}               % <<--- 调整这儿控制列距（最小可设为 0pt）
		% colw: 自动计算使 6*colw + 5*gap = \textwidth
		\newlength{\colw}
		\setlength{\colw}{\dimexpr(\textwidth - 5\gap)/6\relax}
        
		% ================================
		
		% 第一行（6 个）
		\begin{minipage}{\colw}\centering
			% \includesvg[width=\linewidth]{SparseEA2_1}
            \includegraphics[width=\linewidth]{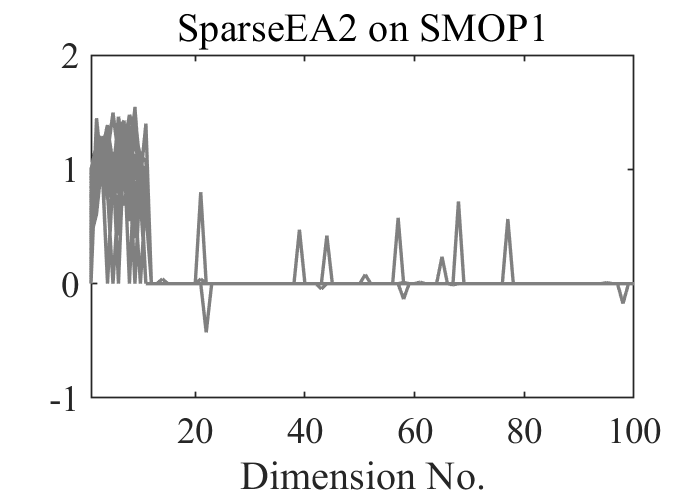}
		\end{minipage}\hspace{\gap}%
		\begin{minipage}{\colw}\centering
			% \includesvg[width=\linewidth]{MSKEA_1}
            \includegraphics[width=\linewidth]{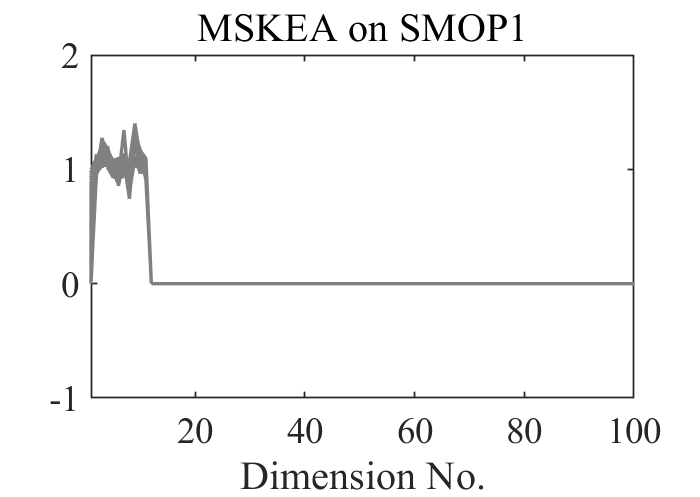}
		\end{minipage}\hspace{\gap}%
		\begin{minipage}{\colw}\centering
			% \includesvg[width=\linewidth]{SNSGAII_1}
	       \includegraphics[width=\linewidth]{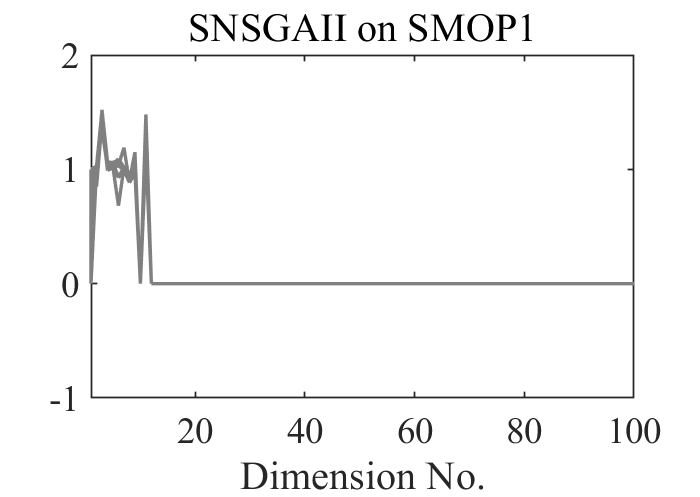}	
        \end{minipage}\hspace{\gap}%
		\begin{minipage}{\colw}\centering
			% \includesvg[width=\linewidth]{TELSO_1}
            \includegraphics[width=\linewidth]{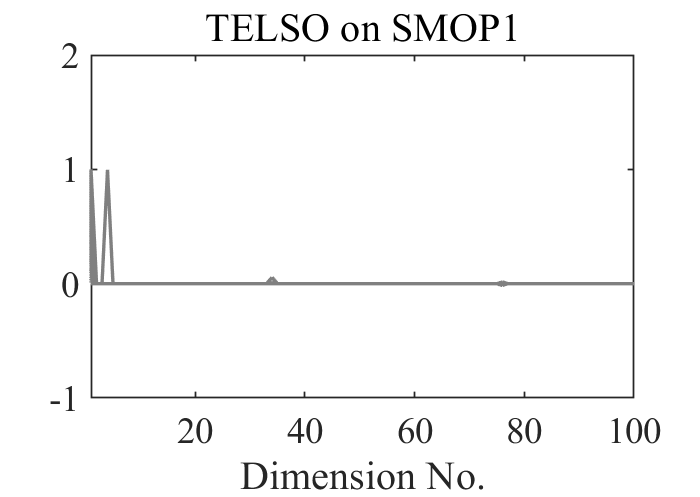}
		\end{minipage}\hspace{\gap}%
		\begin{minipage}{\colw}\centering
			% \includesvg[width=\linewidth]{MGCEA_1}
            \includegraphics[width=\linewidth]{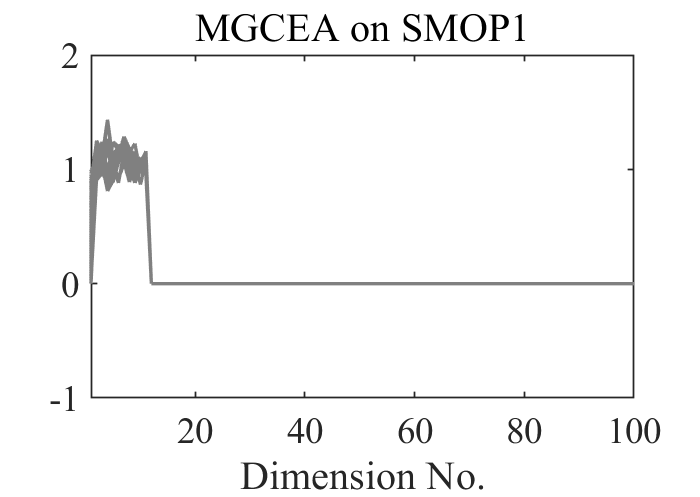}
		\end{minipage}\hspace{\gap}%
		\begin{minipage}{\colw}\centering
			% \includesvg[width=\linewidth]{SMOEA-OPS_1}
            \includegraphics[width=\linewidth]{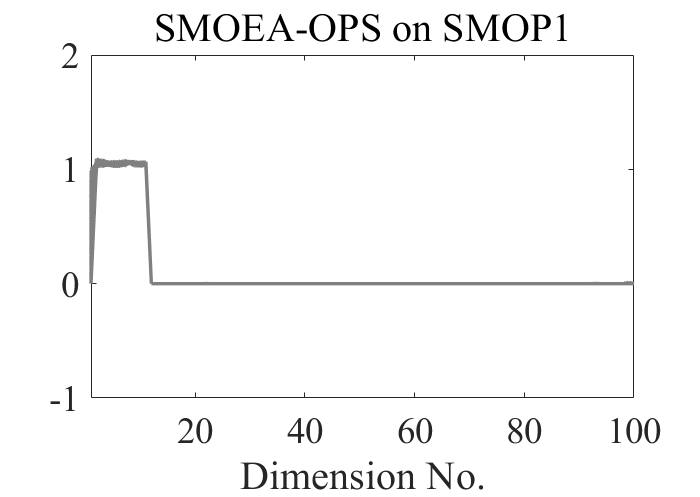}
		\end{minipage}
		
		\vspace{0.4em} % 可按需极小调整行间距
		
		\begin{minipage}{\colw}\centering
			% \includesvg[width=\linewidth]{SparseEA2_1}
            \includegraphics[width=\linewidth]{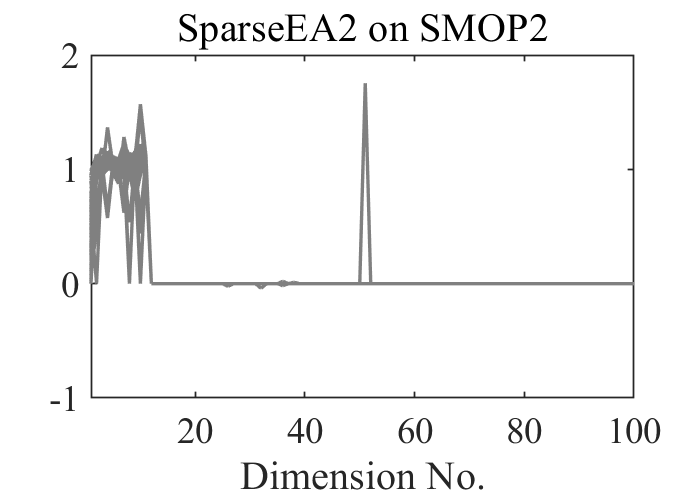}
		\end{minipage}\hspace{\gap}%
		\begin{minipage}{\colw}\centering
			% \includesvg[width=\linewidth]{MSKEA_1}
            \includegraphics[width=\linewidth]{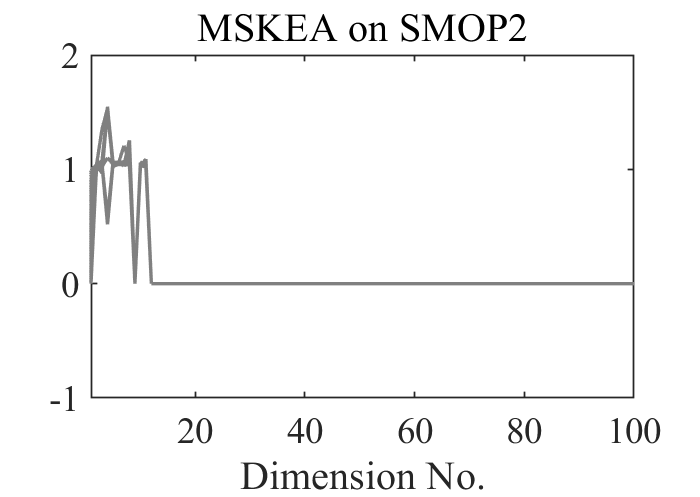}
		\end{minipage}\hspace{\gap}%
		\begin{minipage}{\colw}\centering
			% \includesvg[width=\linewidth]{SNSGAII_1}
	       \includegraphics[width=\linewidth]{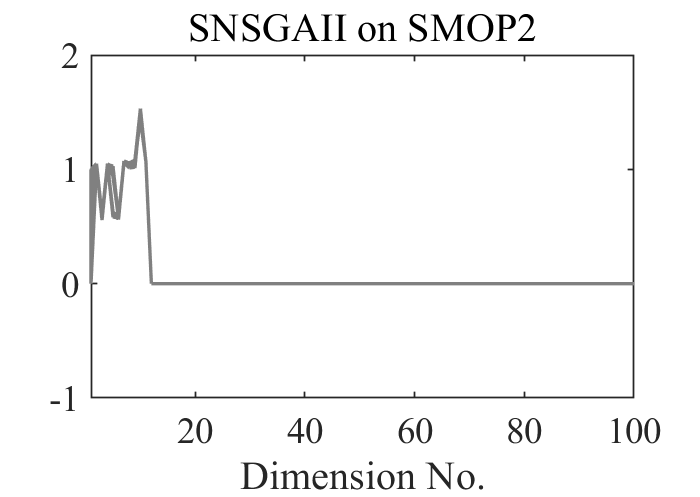}	
        \end{minipage}\hspace{\gap}%
		\begin{minipage}{\colw}\centering
			% \includesvg[width=\linewidth]{TELSO_1}
            \includegraphics[width=\linewidth]{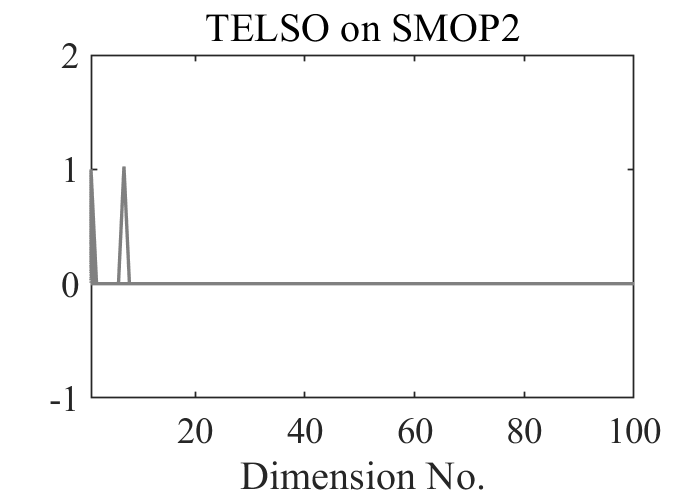}
		\end{minipage}\hspace{\gap}%
		\begin{minipage}{\colw}\centering
			% \includesvg[width=\linewidth]{MGCEA_1}
            \includegraphics[width=\linewidth]{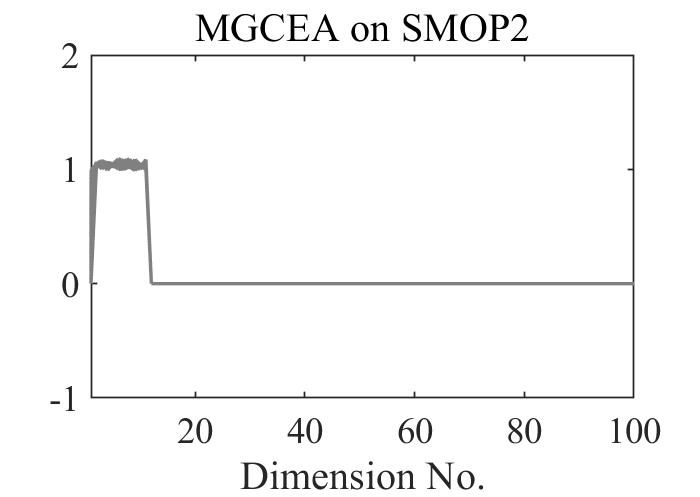}
		\end{minipage}\hspace{\gap}%
		\begin{minipage}{\colw}\centering
			% \includesvg[width=\linewidth]{SMOEA-OPS_1}
            \includegraphics[width=\linewidth]{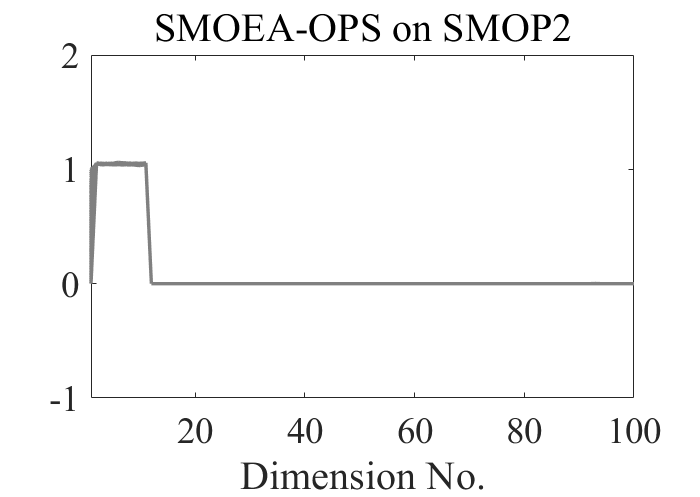}
		\end{minipage}
		\vspace{0.4em}
		 \begin{minipage}{\colw}\centering
			% \includesvg[width=\linewidth]{SparseEA2_1}
            \includegraphics[width=\linewidth]{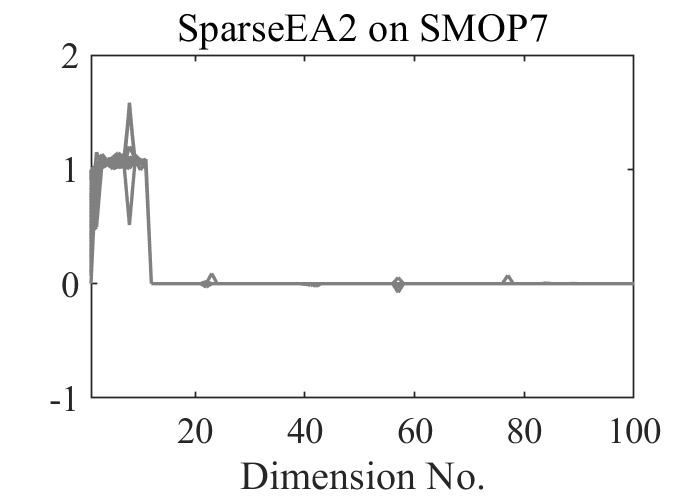}
		\end{minipage}\hspace{\gap}%
		\begin{minipage}{\colw}\centering
			% \includesvg[width=\linewidth]{MSKEA_1}
            \includegraphics[width=\linewidth]{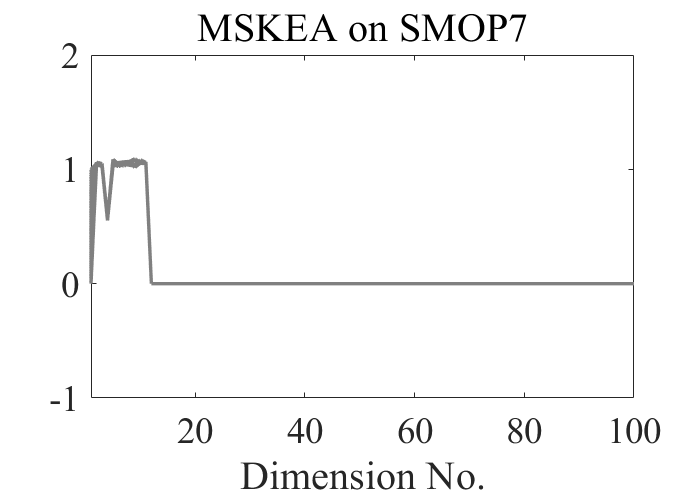}
		\end{minipage}\hspace{\gap}%
		\begin{minipage}{\colw}\centering
			% \includesvg[width=\linewidth]{SNSGAII_1}
	       \includegraphics[width=\linewidth]{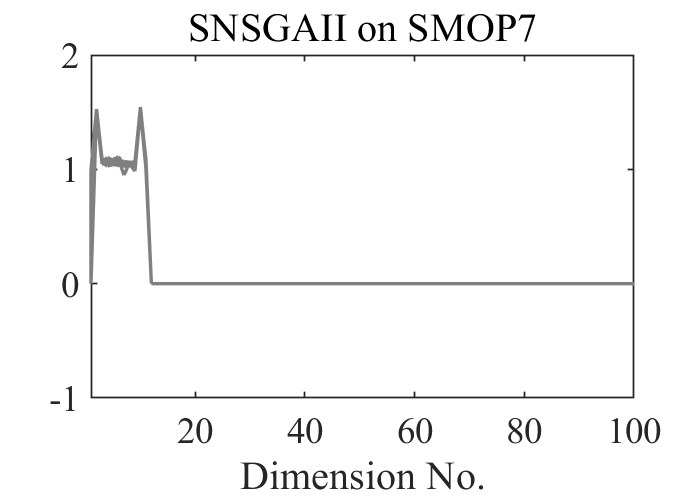}	
        \end{minipage}\hspace{\gap}%
		\begin{minipage}{\colw}\centering
			% \includesvg[width=\linewidth]{TELSO_1}
            \includegraphics[width=\linewidth]{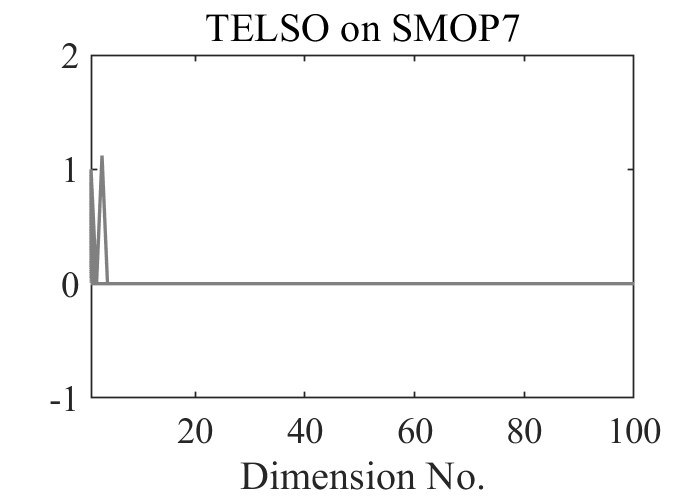}
		\end{minipage}\hspace{\gap}%
		\begin{minipage}{\colw}\centering
			% \includesvg[width=\linewidth]{MGCEA_1}
            \includegraphics[width=\linewidth]{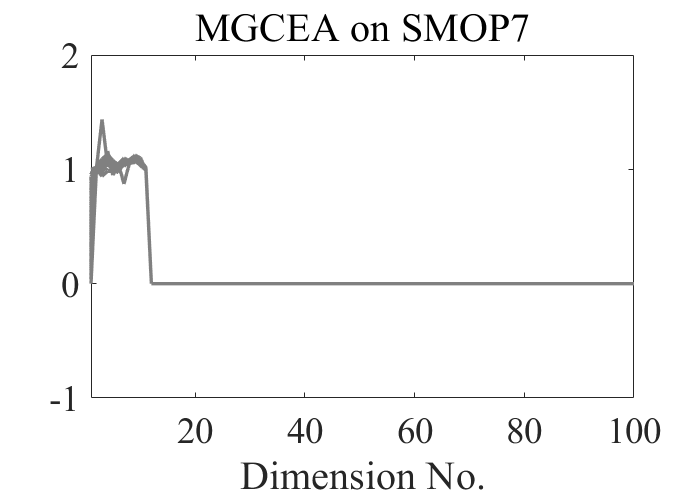}
		\end{minipage}\hspace{\gap}%
		\begin{minipage}{\colw}\centering
			% \includesvg[width=\linewidth]{SMOEA-OPS_1}
            \includegraphics[width=\linewidth]{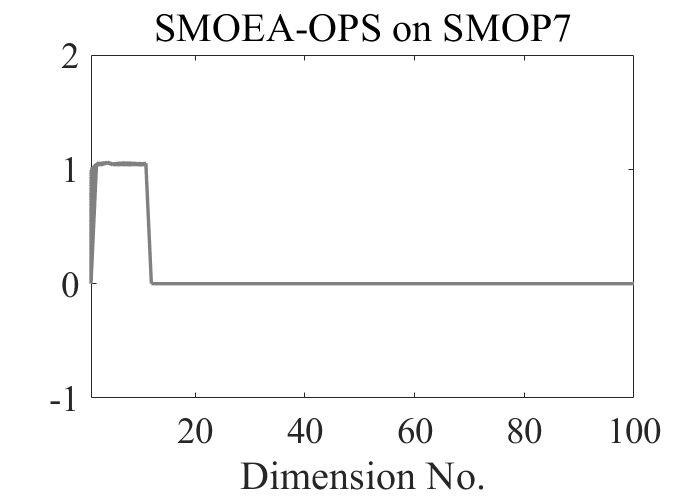}
		\end{minipage}
		\vspace{0.4em}
		\caption{Parallel coordinates of the decision variables obtained by SparseEA2, MSKEA, S-NSGA-II, TELSO, MGCEA, and the proposed SMOEA-OPS on SMOP1, SMOP2, and SMOP7 with 100 decision variables.}
		\label{fig:svg-grid}
	\end{figure*}
	% ----------------------------------------------------------------
     \begin{figure*}[]
	 	\centering
	 	% ========== 调整这里 ==========
	 	% gap: 列间固定间距（可设为 0pt, 0.5pt, 1pt, 0.005\textwidth 等）
	 	%\newlength{\gap}
	 	\setlength{\gap}{-3pt}               % <<--- 调整这儿控制列距（最小可设为 0pt）
	 	% colw: 自动计算使 6*colw + 5*gap = \textwidth
	 	%\newlength{\colw}
	 	\setlength{\colw}{\dimexpr(\textwidth - 2\gap)/3\relax}
	 	% ================================
		
	 	\begin{minipage}{\colw}\centering
	 		% \includesvg[width=\linewidth]{comp1}
                 \includegraphics[width=\linewidth]{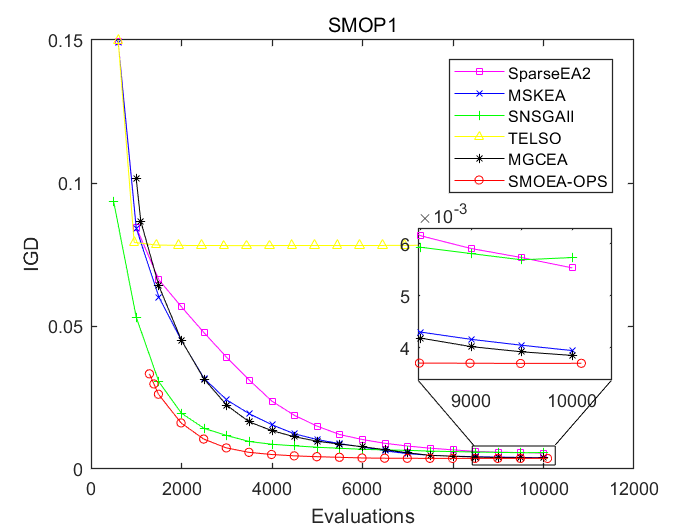}
	 	\end{minipage}\hspace{\gap}%
	 	\begin{minipage}{\colw}\centering
	 		% \includesvg[width=\linewidth]{comp2}
               \includegraphics[width=\linewidth]{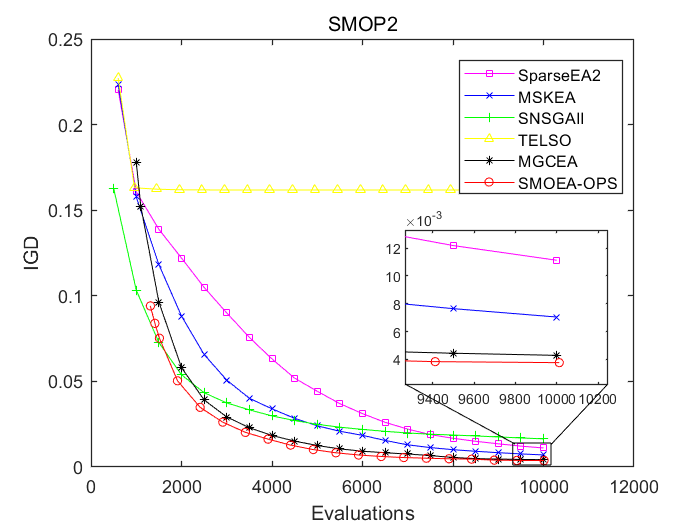}
	 	\end{minipage}\hspace{\gap}%
	 	\begin{minipage}{\colw}\centering
	 		 % \includesvg[width=\linewidth]{comp7}
                 \includegraphics[width=\linewidth]{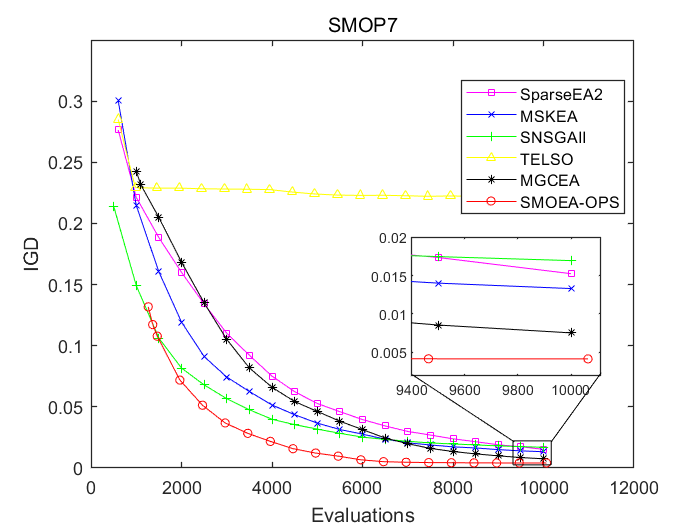}
	 	\end{minipage}\hspace{\gap}%
	 	\caption{Convergence profiles of average IGD values among 30 runs obtained by SparseEA2, MSKEA, S-NSGA-II, TELSO, MGCEA, and the proposed SMOEA-OPS on two-objective SMOP1, SMOP2, and SMOP7 with 100 decision variables.}
	 	\label{fig:svg-grid}
	 \end{figure*}
	
	% ======================== PF图 ============================
	 \begin{figure*}[]
	 	\centering
	 	% 每个图宽度占1/3（减去一点间隙更美观）
	 	\includegraphics[width=0.32\textwidth]{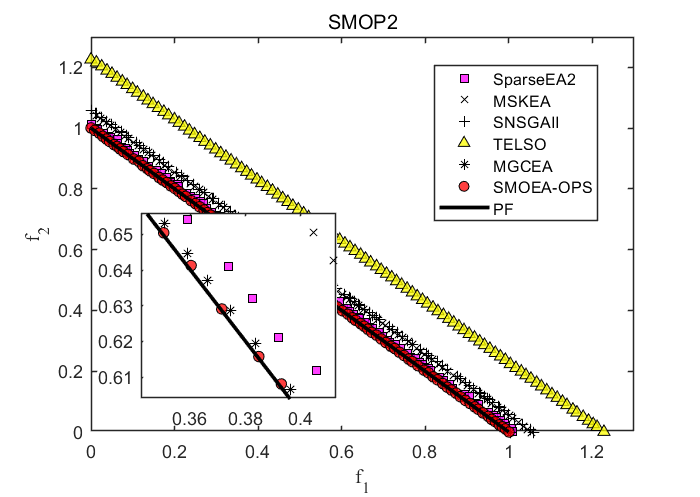}
	 	\includegraphics[width=0.32\textwidth]{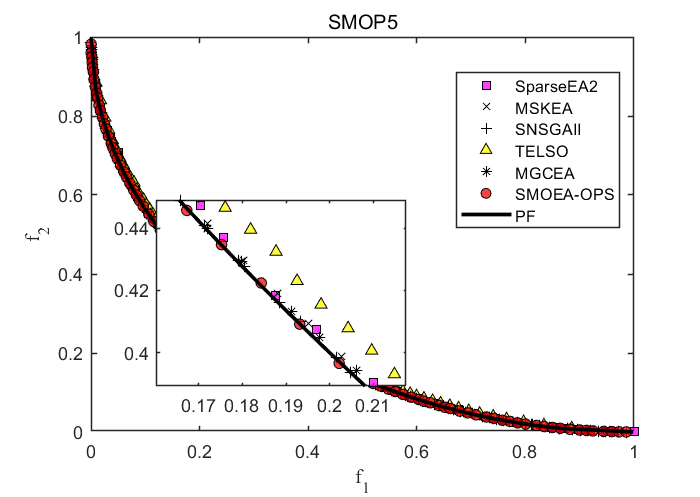}
	 	\includegraphics[width=0.32\textwidth]{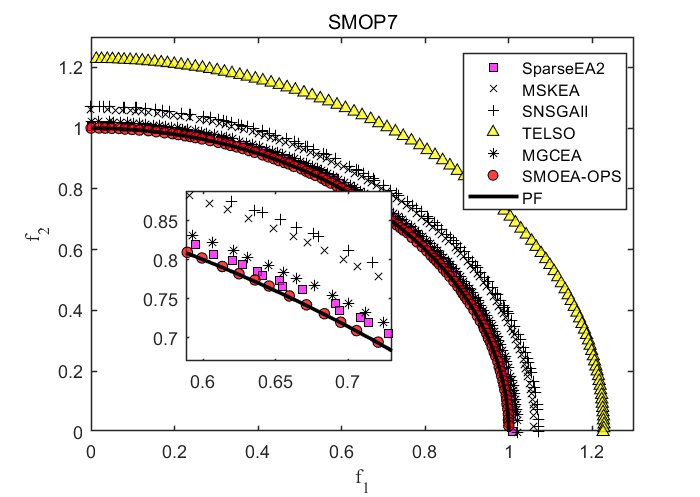}
	 	\caption{Objectives values of the population with average IGD among 30 runs obtained by SparseEA2, MSKEA, S-NSGA-II, TELSO, MGCEA and the proposed SMOEA-OPS on SMOP2, SMOP5 and SMOP7 with 1000 variables.}
	 	\label{fig:three_svgs}
	 \end{figure*}
	% ======================== variables 图 ===================
	\begin{figure*}[]
		\centering
		% 每个图宽度占1/3（减去一点间隙更美观）
		\includegraphics[width=0.24\textwidth]{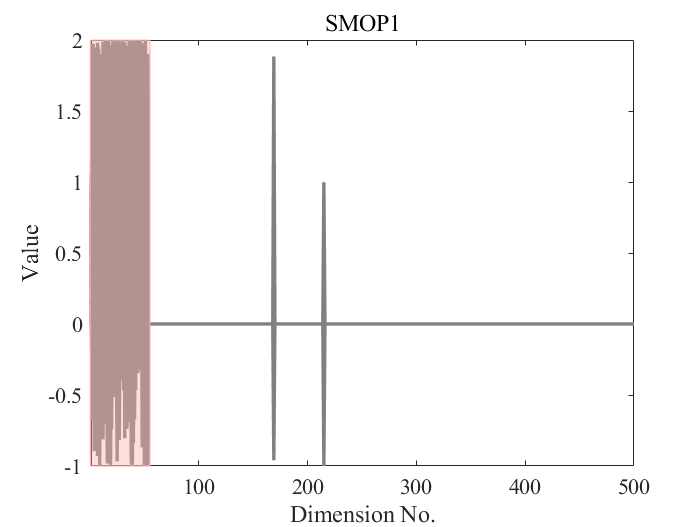}
		\includegraphics[width=0.24\textwidth]{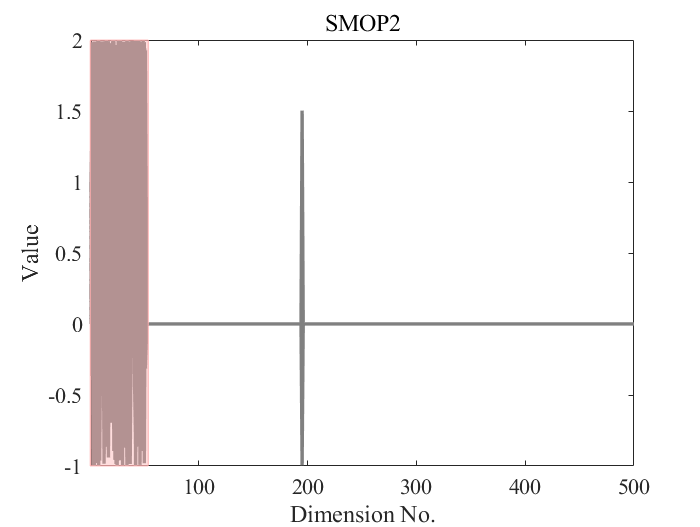}
		\includegraphics[width=0.24\textwidth]{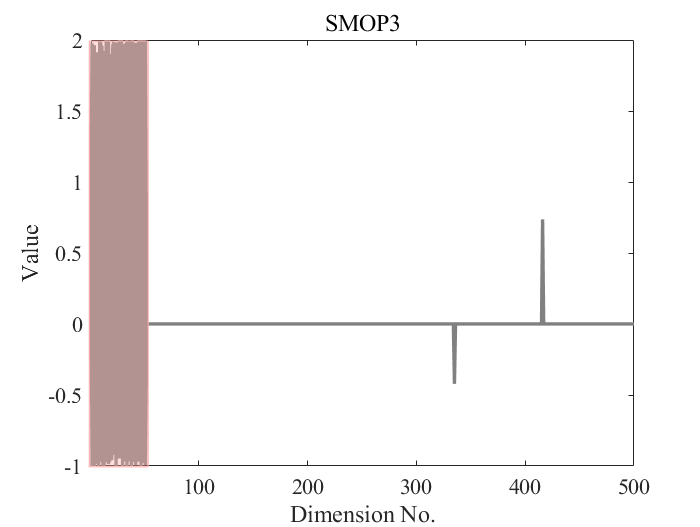}
		\includegraphics[width=0.24\textwidth]{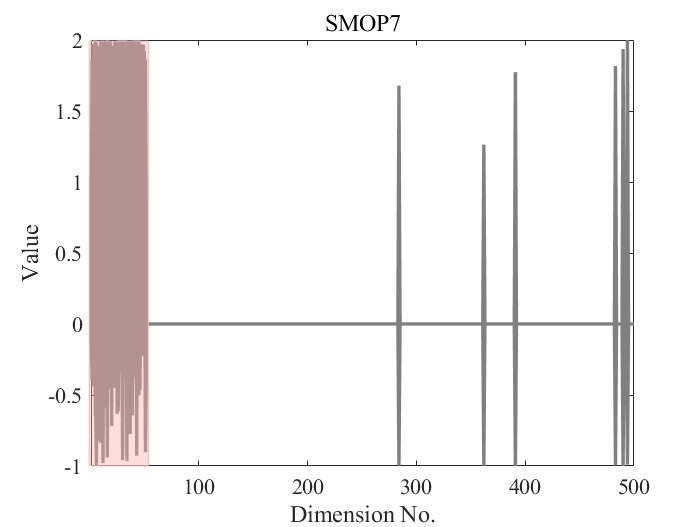}
		\caption{Parallel coordinates of the decision variables obtained by the proposed SMOEA-OPS on SMOP1, SMOP2, SMOP3, and SMOP7 at the 11-th generations. All the variables outside the pink region should be set to zero in the Pareto optimal solutions.}
		\label{fig:three_svgs}
	\end{figure*}
     The optimization results for SMOEA-OPS and the five comparison algorithms are given in Table 2. It is clear that SMOEA-OPS attains the best performance on SMOP1, SMOP2, SMOP3, SMOP5, SMOP6 and SMOP7. Among the 32 test instances, SMOEA-OPS achieves the best results in 22 cases. Specifically, SMOEA-OPS outperforms SparseEA2, MSKEA, S-NAGA-II, TELSO, and MGCEA on 31, 26, 31, 28, and 19 test instances, respectively. \\
    \indent Fig. 5 shows the parallel coordinate plots~\cite{li2017how} of the final decision variables produced by the proposed SMOEA-OPS and the comparison algorithms on SMOP1, SMOP2, and SMOP7. For these problems, Pareto optimal solutions require the first 11 decision variables to be nonzero~\cite{tian2020evolutionary}. It is clear that only SMOEA-OPS and MGCEA accurately identify the nonzero decision variables. SparseEA2 returns solutions that are not sparse enough, while MSKEA, S-NSGA-II, and TELSO produce incomplete solutions -- for example, the final population of S-NSGA-II for SMOP1 includes solutions with the 9-th variable equal to zero. Compared to MGCEA, the final population produced by SMOEA-OPS more accurately approximates the true Pareto front of the SMOPs: nonzero decision variables concentrate tightly around the expected values. For example, on SMOP1, the second through 11-th decision variables in the final solutions of SMOEA-OPS cluster precisely near $\pi/3$~\cite{tian2020evolutionary}, while the corresponding variables in the solutions of MGCEA do not converge to this target. This superior performance is mainly attributed to the Pareto-guided resampling method based on the normal distribution proposed in this paper. These results indicate that SMOEA-OPS can obtain high-quality solutions.\\
    \indent To provide a visual confirmation of SMOEA-OPS’s superiority, Fig. 6 shows the IGD convergence curves of SMOEA-OPS and the compared algorithms on SMOP1, SMOP2, and SMOP7 with 100 decision variables.
    In all three cases, SMOEA-OPS converges much earlier, attaining the IGD values that other algorithms only reach at later stages, and achieves the lowest IGD at the end of the run. Fig. 7 presents the objective values of the population obtained by the six algorithms on SMOP2, SMOP5, and SMOP7 with 1000 decision variables. Across these problems, SMOEA-OPS outperforms the competing algorithms, achieving superior convergence toward the PF while also maintaining greater diversity. Fig. 8 presents the populations obtained at the 5700-th generation, i.e., after 11\% of the evaluation budget (5,700 out of 50,000), from a single run of SMOEA-OPS on SMOP1, SMOP2, SMOP3, and SMOP7 with 500 decision variables. These results show that, even right after initialization, SMOEA-OPS’s population accurately and comprehensively identifies the nonzero decision variables while maintaining the desired sparsity.\\
	\begin{figure}[]
		\centering
        \includegraphics[width=0.5\textwidth]{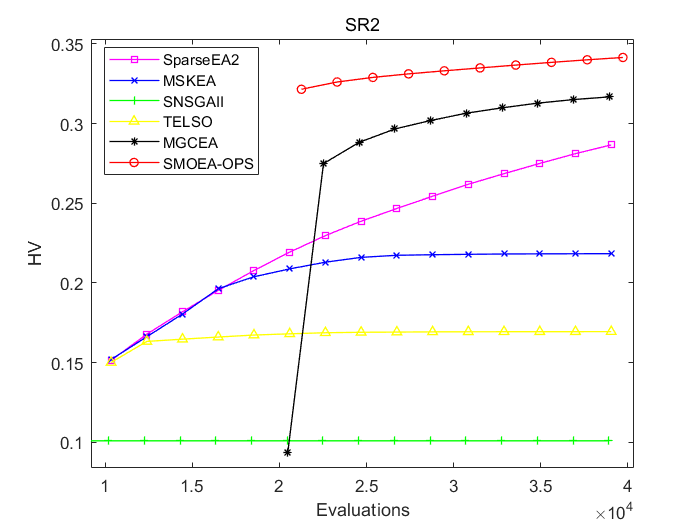} % 不需要.svg后缀
		\caption{Convergence profiles of means of HV values among 30 runs obtained by SparseEA2, MSKEA, S-NSGA-II, TELSO, MGCEA, and the proposed SMOEA-OPS on SR2.}
		\label{fig:svg}
	\end{figure}
    \indent Table 3 reports the mean and standard deviation of HV computed from 30 independent runs for each of the six MOEAs on the practical problems SR1–SR4, KP1–KP4, and CD1–CD4. SMOEA-OPS outperforms all other algorithms on every test instance except CD4, where it performs similarly to MSKEA. In addition, Fig. 9 presents the HV curves on SR2, which show that the population produced by SMOEA-OPS at initialization achieves a higher HV than the final populations of the five competing algorithms. This demonstrates the superiority of the initialization method used in SMOEA-OPS.\\
    \begin{figure}[]
		\centering
		\includegraphics[width=0.5\textwidth]{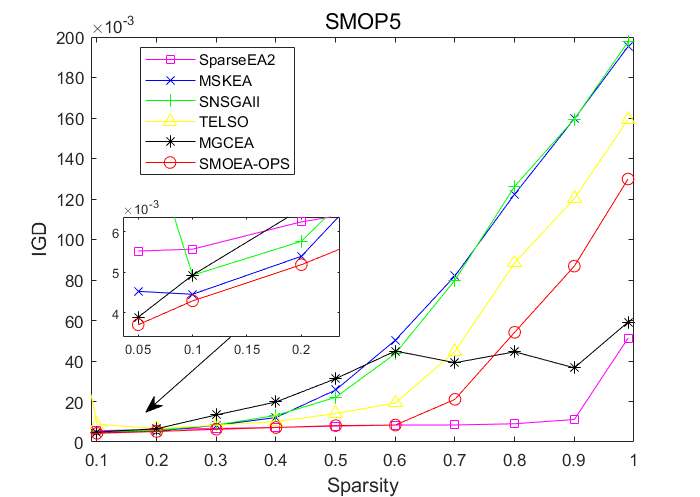} % 不需要.svg后缀
		\caption{IGD values obtained by SparseEA2, MSKEA, S-NAGA-II, TELSO, MGCEA, and the Proposed SMOEA-OPS on SMOP5 with 100 variables, where the sparsity of Pareto optimal solutions is ranged from 0.05 to 1.}
		\label{fig:svg}
	\end{figure}
    \indent Fig. 10 plots the average IGD curves of six algorithms among 30 runs on SMOP5 with 100 decision variables as the sparsity parameter varies from 0.05 to 1. The results show that when the sparsity is below \(\,0.7\,\), SMOEA-OPS achieves lower IGD than the other five algorithms. Overall, the results indicate that SMOEA-OPS performs better on sparser problems and remains reasonably robust across different sparsity levels.
	\subsection{Ablation studies}
	% =============================== 消融实验 ===========================
    \begin{table}[]
		\centering
		\renewcommand{\arraystretch}{1.2}
		\setlength{\tabcolsep}{5pt} % 进一步减小列间距
        \caption{mean of IGD values among 30 runs obtained by SMOEA-OPS, SMOEA-OPS', SMOEA-OPS'', and SMOEA-OPS''' on SMOP1–SMOP8 with 100 decision variables.}
		{ % 使用更小的字体
        \scriptsize
			\begin{tabular}{c *{4}{c}}
				\toprule
				\textbf{Problem} & \textbf{SMOEA-OPS\('\)} & \textbf{SMOEA-OPS\(''\)} & \textbf{SMOEA-OPS\('''\)} & \textbf{SMOEA-OPS} \\
				\midrule
				SMOP1 & 3.7934e-3 $-$ & 3.7159e-3 $=$ & 3.7022e-3 $=$ & \cellcolor{lightgray}3.6989e-3 \\
				SMOP2 & 4.4501e-3 $-$ & 5.1944e-3 $-$ & 3.7370e-3 $-$ & \cellcolor{lightgray}3.7031e-3 \\
				SMOP3 & 3.8297e-3 $-$ & 3.7028e-3 $=$ & 3.7154e-3 $-$ & \cellcolor{lightgray}3.7002e-3 \\
				SMOP4 & 4.9342e-3 $=$ & 4.9341e-3 $=$ & 5.2911e-3 $=$ & \cellcolor{lightgray}4.8039e-3 \\
				SMOP5 & 4.5175e-3 $-$ & 4.3574e-3 $-$ & 4.6222e-3 $-$ & \cellcolor{lightgray}4.2897e-3 \\
				SMOP6 & 5.1227e-3 $-$ & 6.0448e-3 $-$ & 4.6007e-3 $=$ & \cellcolor{lightgray}4.5788e-3 \\
				SMOP7 & 4.8326e-3 $-$ & 2.1510e-2 $-$ & 4.1738e-3 $-$ & \cellcolor{lightgray}4.1308e-3 \\
				SMOP8 & 7.2436e-2 $=$ & 7.8175e-2 $=$ & \cellcolor{lightgray} 6.4134e-2 $=$ & 7.4686e-2 \\
				
				\addlinespace[1pt]
				\bottomrule
				\multirow{1}{*}{\centering + / - / $\approx$}
				& 0/6/2 & 0/4/4 & 0/4/4 \\
				\midrule
			\end{tabular}
		}
		
		\label{fig:abatable}
	\end{table}
	
	\begin{figure}[]
		\centering
      \begin{minipage}[t]{0.49\textwidth}
        \centering
        \includegraphics[width=\linewidth]{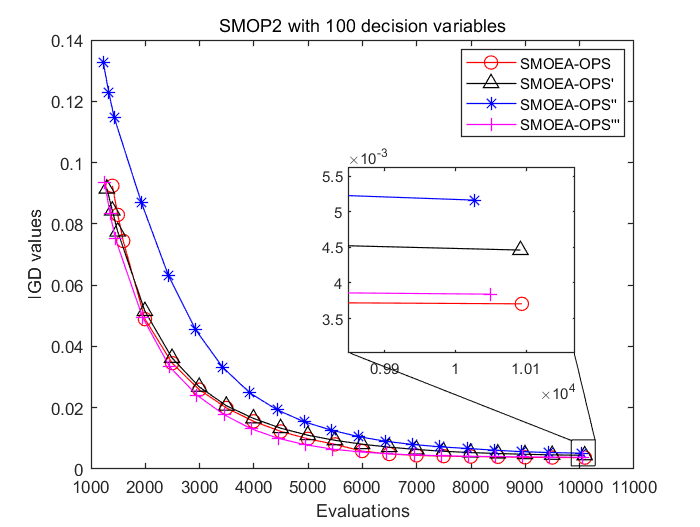}
      \end{minipage}\hfill
      \begin{minipage}[t]{0.49\textwidth}
        \centering
        \includegraphics[width=\linewidth]{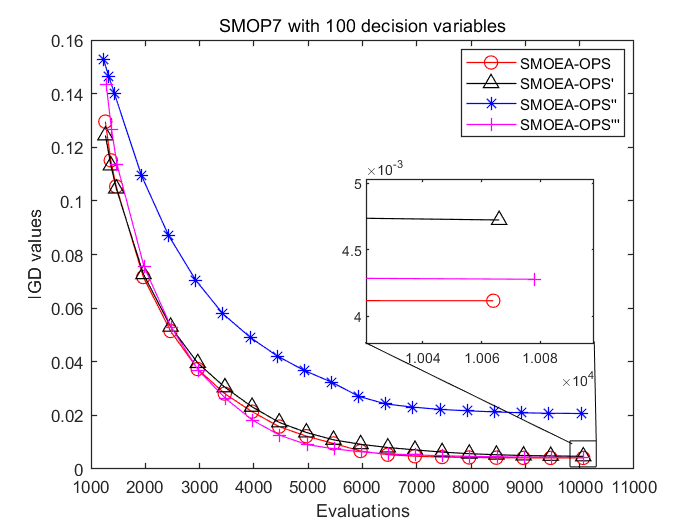}
      \end{minipage}
		\caption{Convergence profiles of means of IGD values among 30 runs obtained by SMOEA-OPS, SMOEA-OPS', SMOEA-OPS'', SMOEA-OPS''' on SMOP2 and SMOP7 with 100 variables.}
		
		\label{fig:svg}
	\end{figure}
	% -------------------------------------------------------------------
% \begin{figure}[htbp]
%     \centering
%     \begin{minipage}{0.48\textwidth}
%         \centering
%         \includegraphics[width=\linewidth]{figs/SMOP2_100_aba.png}
%         \caption*{SMOP2} % 可选的小标题，不编号
%     \end{minipage}
%     \hfill
%     \begin{minipage}{0.48\textwidth}
%         \centering
%         \includegraphics[width=\linewidth]{figs/SMOP7_100_aba.png}
%         \caption*{SMOP7} % 可选的小标题，不编号
%     \end{minipage}
%     \caption{Convergence profiles of means of IGD values among 30 runs obtained by SMOEA-OPS, SMOEA-OPS', SMOEA-OPS'', SMOEA-OPS''' on SMOP2 and SMOP7 with 100 variables.}
%     \label{fig:svg}
% \end{figure}
	SMOEA-OPS's superior performance is mainly attributed to three components: the Pareto-guided resampling method based on the normal distribution for optimizing real vectors, the new score of variable computation, and the dynamic mutation probabilities assigned to each real variable for different solutions. To demonstrate the contribution of these three components to the optimization results, three variants of SMOEA-OPS are proposed for ablation experiments, namely SMOEA-OPS\('\), SMOEA-OPS\(''\), and SMOEA-OPS\('''\). SMOEA-OPS\('\) is obtained by removing the Pareto-guided resampling in SMOEA-OPS. SMOEA-OPS\(''\) is obtained by replacing the \(Score\) with the sum of the non-dominated front numbers of the \(NVal\) intervals in SMOEA-OPS, which is the score calculated by the MGCEA method. SMOEA-OPS\('''\) is obtained by replacing \(mutProb\) with $1/D$ as the variance probability of real variables.\\
    \indent Table 4 shows the mean and standard deviation of the IGD obtained by SMOEA-OPS and variants of SMOEA-OPS on the SMOP benchmark problem with 100 decision variables. The deterioration of the optimization results can be observed through the three variants of SMOEA-OPS due to the absence of these three components.\\
	\indent Fig. 11 visually illustrates the convergence profiles in terms of IGD values for SMOEA-OPS and its three variants. SMOEA-OPS achieves a faster decrease in IGD than SMOEA-OPS\('\) and SMOEA-OPS\(''\). SMOEA-OPS\('''\) achieves a faster decrease in IGD than SMOEA-OPS, but its final solutions are inferior. This is because SMOEA-OPS's \(mutProb\) encourages stronger exploration in the early stages, enabling the algorithm to escape local optima and obtain better final solutions. Overall, SMOEA-OPS\('\) performs worst, which shows the importance of the proposed Pareto-guided resampling strategy. On SMOP7, SMOEA-OPS\(''\) records a much higher IGD than the others, indicating that the proposed calculation of score is effective on high-difficulty problems with both epistasis and multimodal landscapes~\cite{shao2025knowledge}.
    \subsection{Sensitivity analysis}
	The vector \(Score\) is calculated based on \(SVal\) and \(NVal\) like MGCEA. Fig. 12 shows a heatmap of the total IGD across SMOP1–SMOP8 for varying NVal and SVal, with 100 decision variables and 2 objectives. Note that to avoid the IGD of SMOP8 dominating the results, divide the SMOP8 value by 10 in advance. We can conclude from Fig. 12 that when \(NVal\) and \(SVal\) take 5 and 2, respectively, the sum of IGD across SMOP1-SMOP8 is the lowest, that is, the algorithm performs the best. If \(SVal\) and \(NVal\) are further increased, it will result in more evaluations being consumed. Therefore, \(NVal\) and \(SVal\) are set to 5 and 2, respectively.\\

    \begin{figure}[]
		\centering
		\includegraphics[width=0.5\textwidth]{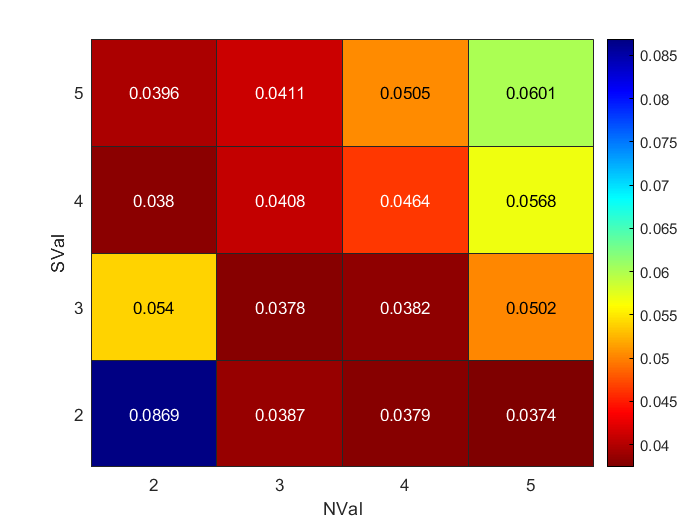} % 不需要.svg后缀
		\caption{Heatmap of the sum of IGD values across SMOP1-SMOP8 as \(NVal\) and \(SVal\) vary from 2 to 5.}
		\label{fig:svg}
	\end{figure}

	% ========================= runtime ==================
	\begin{table}[]
		\centering
		\renewcommand{\arraystretch}{1.1}
		\setlength{\tabcolsep}{2pt}
        \caption{Means of runtime (in second) among 30 runs obtained by SparseEA2, MSKEA, S-NSGA-II, TELSO, MGCEA, and the proposed SMOEA-OPS on SMOP1–SMOP8.}
		{\scriptsize
			\begin{tabular}{c c *{6}{c}}
				\toprule
				\textbf{Problem} & \textbf{D} & \textbf{SparseEA2} & \textbf{MSKEA} & \textbf{SNSGAII} & \textbf{TELSO} & \textbf{MGCEA} & \textbf{SMOEA-OPS} \\
				\midrule
				\multirow{4}{*}{\makecell{SMOP1-8\\(average)}} 
				& 100   & 8.61e-1    & 9.33e-1    & \cellcolor{lightgray}3.18e-1    & 7.81e-1    & 1.13e+0    & 1.36e+0    \\
				& 500   & 9.04e+0    & 8.10e+0    & \cellcolor{lightgray}2.74e+0    & 5.35e+0    & 1.05e+1    & 9.78e+0    \\
				& 1000  & 3.38e+1    & 2.74e+1    &\cellcolor{lightgray} 9.59e+0    & 1.88e+1    & 3.36e+1    & 2.94e+1    \\
				& 5000  & 8.36e+2    & 5.74e+2    &\cellcolor{lightgray} 2.26e+2    & 5.27e+2    & 6.33e+2    & 6.75e+2    \\
				\midrule
				SR2    & 2048  & 3.02e+1    & 2.29e+1    &\cellcolor{lightgray} 1.49e+1    & 2.04e+1    & 2.23e+1    & 2.84e+1    \\
				\midrule
				KP4    & 5000  &\cellcolor{lightgray} 1.46e+2    & 1.68e+2    & 1.99e+2    & 2.82e+2    & 2.22e+2    & 1.75e+2    \\
				\midrule
				CD4    & 115   & 4.76e+0    & 5.44e+0    &\cellcolor{lightgray} 3.34e+0    & 4.70e+0    & 5.36e+0    & 6.04e+0    \\
				
				\bottomrule
				\addlinespace[4pt]
			\end{tabular}
		}
		\label{tab:runtime_results}
	\end{table}
	\subsection{Computational efficiency}
	Table 5 shows a comparison of the runtimes of the proposed SMOEA-OPS and the other five algorithms. The results show that the proposed SMOEA-OPS is competitive with MGCEA and SparseEA2. For the higher-dimensional KP4 problem, whose solutions are represented as binary vectors, SMOEA-OPS has a runtime that is outperformed only by SparseEA2 and MSKEA due to its use of relatively simple genetic operators. In summary, S-NSGA-II has the shortest runtime on most problems, followed by TELSO. The runtimes of the other four algorithms are comparable. However, S-NSGA-II and TELSO yield the worst optimization results, whereas SMOEA-OPS obtains the best results within a reasonable running time.
\section{Conclusions}
	LSSMOPs are still very challenging, mainly because most LSSMOOAs do not obtain populations that can accurately cover nonzero variables after initialization. Consequently, the population typically requires extended evolution or a large number of expensive function evaluations to discover the relevant variables. This paper proposes A large-scale sparse multiobjective optimization algorithm based on optimal performance scores. 
    Specifically, this paper proposes three key techniques: a new initialization strategy, a dynamic mutation-probability calculation based on the performance of real variables across intervals, and a Pareto-guided resampling method based on the normal distribution for optimizing real vectors. The proposed initialization method produces initial populations that accurately cover the nonzero variables. The proposed mutation probability and Pareto-guided resampling improve real vector optimization and enable the algorithm to escape local optima. These three components can be easily embedded into other algorithms. Finally, we validate SMOEA-OPS on a suite of eight benchmark SMOPs and three real-world SMOPs; the results show that SMOEA-OPS is consistently effective and achieves superior performance. \\
    \indent In future work, we aim to further reduce the function evaluation cost incurred during initialization, thereby preserving a larger share of the evaluation budget for subsequent evolutionary search. We will replace the manual parameters \(NVal\) and \(SVal\) with an adaptive, problem-driven computation. Moreover, we will explore more efficient methods for optimizing real variables in LSSMOPs to mitigate the computational overhead introduced by resampling.

\section*{CRediT authorship contribution statement}
\textbf{Jia-Lin Mai}: Writing - review \& editing, Writing - original draft, 
Software, Conceptualization, Formal analysis.
\textbf{Min-Rong Chen}: Methodology, Writing - original draft, Writing - review \& editing, Funding acquisition. \textbf{Guo-Qiang Zeng}: Supervision, Funding acquisition. \textbf{Xiang Liu}: Supervision, Writing - review \& editing, Funding acquisition. \textbf{Jian Weng}: Project administration, Funding acquisition. 
\section*{Declaration of competing interest}
The authors declare that they have no known competing financial interests or personal relationships that could have appeared to influence the work reported in this paper.
\section*{Acknowledgement}
This work was supported in part by the National Natural Science Foundation of China (Grant Nos. 62573201, U23A20303, 62573326, 62533016, 61825203, 62332007, and U22B2028), in part by the Zhejiang Provincial Natural Science Foundation of China (Grant No. LZ25F030007), in part by the Outstanding Youth Project of Guangdong Basic and Applied Basic Research Foundation (Grant No. 2023B1515020064), and in part by the Guangdong Basic and Applied Basic Research Foundation (Grant No：2024A1515140109).
\bibliographystyle{unsrt}
\bibliography{references}

\begin{thebibliography}{10}

\bibitem{coello2006evolutionary}
CA~Coello Coello.
\newblock Evolutionary multi-objective optimization: a historical view of the
  field.
\newblock {\em IEEE computational intelligence magazine}, 1(1):28--36, 2006.

\bibitem{machinelearning}
Y.~Jin, T.~Okabe, and B.~Sendhoff.
\newblock Evolutionary multi-objective optimization approach to constructing
  neural network ensembles for regression.
\newblock In C.~A. Coello and Gilles Lamont, editors, {\em Applications of
  Multi-Objective Evolutionary Algorithms}, pages 653--672. World Scientific,
  Hackensack, NJ, USA, 2004.

\bibitem{zhou2011multiobjective}
A.~Zhou, B.~Y. Qu, H.~Li, S.~Zhao, P.~N. Suganthan, and Q.~Zhang.
\newblock Multiobjective evolutionary algorithms: A survey of the state of the
  art.
\newblock {\em Swarm and evolutionary computation}, 1(1):32--49, 2011.

\bibitem{tang2012incorporating}
L.~Tang, L.~Zhang, P.~Luo, and M.~Wang.
\newblock Incorporating occupancy into frequent pattern mining for high quality
  pattern recommendation.
\newblock In {\em Proceedings of the 21st ACM International Conference on
  Information and Knowledge Management}, pages 75--84. ACM, 2012.

\bibitem{economics}
A.~Ponsich, A.~L. Jaimes, and C.~A.~Coello Coello.
\newblock A survey on multiobjective evolutionary algorithms for the solution
  of the portfolio optimization problem and other finance and economics
  applications.
\newblock {\em IEEE Trans. on Evolutionary Computation}, 17(3):321--344, 2013.

\bibitem{qi2024enhancing}
S.~Qi, R.~Wang, T.~Zhang, W.~Huang, F.~Yu, and L.~Wang.
\newblock Enhancing evolutionary algorithms with pattern mining for sparse
  large-scale multi-objective optimization problems.
\newblock {\em IEEE-CAA J. Automatica Sinica}, 11(8):1786--1801, 2024.

\bibitem{xue2013particle}
B.~Xue, M.~Zhang, and W.~N. Browne.
\newblock Particle swarm optimization for feature selection in classification:
  A multi-objective approach.
\newblock {\em IEEE Trans. on Cybernetics}, 43(6):1656--1671, December 2013.

\bibitem{li2018preference}
H.~Li, Q.~Zhang, J.~Deng, and Z.-B. Xu.
\newblock A preference-based multiobjective evolutionary approach for sparse
  optimization.
\newblock {\em IEEE Trans. on Neural Networks and Learning Systems},
  29(5):1716--1731, May 2018.

\bibitem{fieldsend2005pareto}
J.~E. Fieldsend and S.~Singh.
\newblock Pareto evolutionary neural networks.
\newblock {\em IEEE Trans. on Neural Networks and Learning Systems},
  16(2):338--354, March 2005.

\bibitem{tan2021multi}
Z.~Tan, H.~Wang, and S.~Liu.
\newblock Multi-stage dimension reduction for expensive sparse multi-objective
  optimization problems.
\newblock {\em Neurocomputing}, 440:159--174, 2021.

\bibitem{shao2025evolutionary}
S.~Shao, Y.~Tian, Y.~Zhang, S.~Yang, P.~Zhang, C.~He, X.~Zhang, and Y.~Jin.
\newblock Evolutionary computation for sparse multi-objective optimization: A
  survey.
\newblock {\em ACM Computing Surveys}, 57(11):1--35, 2025.

\bibitem{yang2025sparse}
W.~Yang, J.~Liu, Y.~Liu, and T.~Zheng.
\newblock A sparse large-scale multi-objective evolutionary algorithm based on
  sparsity detection.
\newblock {\em Swarm and Evolutionary Computation}, 92:101820, 2025.

\bibitem{huang2025enhanced}
X.~Huang, J.~Wang, K.~Zhang, B.~Yuan, C.~Dai, and S.~V. Ablameyko.
\newblock An enhanced sparse multiobjective evolutionary algorithm in
  large-scale multiobjective optimization.
\newblock {\em Information Sciences}, page 122476, 2025.

\bibitem{zhou2024highdim}
M.~C. Zhou, M.~Cui, D.~Xu, S.~Zhu, Z.~Zhao, and A.~Abusorrah.
\newblock Evolutionary optimization methods for high-dimensional expensive
  problems: A survey.
\newblock {\em IEEE-CAA J. Automatica Sinica}, 11(5):1092--1105, 2024.

\bibitem{hua2021irregularpf}
Y.~Hua, Q.~Liu, K.~Hao, and Y.~Jin.
\newblock A survey of evolutionary algorithms for multi-objective optimization
  problems with irregular pareto fronts.
\newblock {\em IEEE-CAA J. Automatica Sinica}, 8(2):303--318, 2021.

\bibitem{7544478}
X.~Zhang, Y.~Tian, R.~Cheng, and Y.~Jin.
\newblock A decision variable clustering-based evolutionary algorithm for
  large-scale many-objective optimization.
\newblock {\em IEEE Transactions on Evolutionary Computation}, 22(1):97--112,
  2018.

\bibitem{jiang2025heterogeneous}
J.~Jiang, H.~Wang, P.~Tong, J.~Hong, Z.~Liu, X.~Zhuang, B.~Su, and F.~Han.
\newblock A heterogeneous sparsity knowledge guided evolutionary algorithm for
  sparse large-scale multiobjective optimization.
\newblock {\em Swarm and Evolutionary Computation}, 96:102000, 2025.

\bibitem{guo2025constrained}
J.~Guo and Y.~Shan.
\newblock A constrained multi-objective evolutionary algorithm with weak
  constraint--pareto dominance and angle distance-based diversity preservation.
\newblock {\em Mathematics}, 13(22):3696, 2025.

\bibitem{wang2025cmoea}
Y.~Wang and H.~Liu.
\newblock A constrained multi-objective evolutionary algorithm with weak
  constraint–pareto dominance.
\newblock {\em Mathematics}, 13(22):3696, 2025.

\bibitem{A_CO_E}
Y.~Zhang, C.~Wu, Y.~Tian, and X.~Zhang.
\newblock A co-evolutionary algorithm based on sparsity clustering for sparse
  large-scale multi-objective optimization.
\newblock {\em Engineering Applications of Artificial Intelligence}, 133(PB),
  July 2024.

\bibitem{jiang2023two}
J.~Jiang, F.~Han, J.~Wang, Q.~Ling, H.~Han, and Y.~Wang.
\newblock A two-stage evolutionary algorithm for large-scale sparse
  multiobjective optimization problems.
\newblock {\em Swarm and Evolutionary Computation}, 70:101060, 2023.

\bibitem{ye2022solving}
Y.~Tian, C.~Lu, X.~Zhang, K.~C. Tan, and Y.~Jin.
\newblock Solving large-scale multiobjective optimization problems with sparse
  optimal solutions via unsupervised neural networks.
\newblock {\em IEEE Trans. on Cybernetics}, 52(99):1--13, 2022.

\bibitem{10477568}
L.~Pan, J.~Lin, H.~Wang, C.~He, K.~C. Tan, and Y.~Jin.
\newblock Computationally expensive high-dimensional multiobjective
  optimization via surrogate-assisted reformulation and decomposition.
\newblock {\em IEEE Transactions on Evolutionary Computation}, 29(4):921--935,
  2025.

\bibitem{ye2024mskea}
Z.~Ding, L.~Chen, D.~Sun, and X.~Zhang.
\newblock A multi-stage knowledge-guided evolutionary algorithm for large-scale
  sparse multi-objective optimization problems.
\newblock {\em Swarm and Evolutionary Computation}, 84:101456, 2024.

\bibitem{wang2025evolution}
X.~Wang, W.~Zhao, J.~Tang, Z.~Dai, and Y.~Feng.
\newblock Evolution algorithm with adaptive genetic operator and dynamic
  scoring mechanism for large-scale sparse many-objective optimization.
\newblock {\em Scientific Reports}, 15(1):9267, 2025.

\bibitem{liu2021variable}
S.~Liu, Q.~Lin, Y.~Tian, and K.~C. Tan.
\newblock A variable importance-based differential evolution for large-scale
  multiobjective optimization.
\newblock {\em IEEE Trans. on Cybernetics}, 52(12):13048--13062, 2021.

\bibitem{li2025surrogate}
W.~Li, Y.~Qiu, Z.~Wang, B.~Xu, Z.~Hao, Q.~Zhang, Y.~Li, and Z.~Fan.
\newblock Surrogate-assisted neural learning and evolutionary optimization for
  expensive constrained multi-objective problems.
\newblock {\em Swarm and Evolutionary Computation}, 97:102020, 2025.

\bibitem{liang2024multi}
J.~Liang, K.~Zhu, Y.~Li, Y.~Li, and Y.~Gong.
\newblock Multi-objective evolutionary neural architecture search with
  weight-sharing supernet.
\newblock {\em Applied Sciences}, 14(14):6143, 2024.

\bibitem{santos2023neuroevolution}
Frederico~JJB Santos, Ivo Gon{\c{c}}alves, and Mauro Castelli.
\newblock Neuroevolution with box mutation: An adaptive and modular framework
  for evolving deep neural networks.
\newblock {\em Applied Soft Computing}, 147:110767, 2023.

\bibitem{marler2004survey}
R.~T. Marler and J.~S. Arora.
\newblock Survey of multi-objective optimization methods for engineering.
\newblock {\em Structural and multidisciplinary optimization}, 26(6):369--395,
  2004.

\bibitem{tian2021evolutionary}
Y.~Tian, L.~Si, X.~Zhang, R.~Cheng, C.~He, K.~C. Tan, and Y.~Jin.
\newblock Evolutionary large-scale multi-objective optimization: A survey.
\newblock {\em ACM Computing Surveys}, 54(8):1--34, 2021.

\bibitem{wang2024sparse}
X.~Wang, R.~Cheng, and Y.~Jin.
\newblock Sparse large-scale multiobjective optimization by identifying nonzero
  decision variables.
\newblock {\em IEEE Trans. on Systems, Man and Cybernetics: Systems},
  54(10):1--13, 2024.

\bibitem{mckay2000comparison}
M.~D. McKay, R.~J. Beckman, and W.~J. Conover.
\newblock A comparison of three methods for selecting values of input variables
  in the analysis of output from a computer code.
\newblock {\em Technometrics}, 42(1):55--61, 2000.

\bibitem{wang2022enhanced}
X.~Wang, K.~Zhang, J.~Wang, and Y.~Jin.
\newblock An enhanced competitive swarm optimizer with strongly convex sparse
  operator for large-scale multiobjective optimization.
\newblock {\em IEEE Trans. on Evolutionary Computation}, 26(5):859--871,
  October 2022.

\bibitem{tian2022pattern}
Y.~Tian, C.~Lu, X.~Zhang, F.~Cheng, and Y.~Jin.
\newblock A pattern mining-based evolutionary algorithm for large-scale sparse
  multiobjective optimization problems.
\newblock {\em IEEE Trans. on Cybernetics}, 52(7):6784--6797, July 2022.

\bibitem{tian2020evolutionary}
Y.~Tian, X.~Zhang, C.~Wang, and Y.~Jin.
\newblock An evolutionary algorithm for large-scale sparse multiobjective
  optimization problems.
\newblock {\em IEEE Trans. on Evolutionary Computation}, 24(2):380--393, April
  2020.

\bibitem{MGCEA}
Y.~Tian, S.~Shao, G.~Xie, and X.~Zhang.
\newblock A multi-granularity clustering based evolutionary algorithm for
  large-scale sparse multi-objective optimization.
\newblock {\em Swarm and Evolutionary Computation}, 84:101453, 2024.

\bibitem{shao2025knowledge}
S.~Shao, Y.~Tian, Y.~Zhang, and X.~Zhang.
\newblock Knowledge learning-based dimensionality reduction for solving
  large-scale sparse multiobjective optimization problems.
\newblock {\em IEEE Trans. on Cybernetics}, page PP(99), April 2025.

\bibitem{SparseEA2}
Y.~Zhang, Y.~Tian, and X.~Zhang.
\newblock Improved sparseea for sparse large-scale multi-objective optimization
  problems.
\newblock {\em Complex \& Intelligent Systems}, 9:1127--1142, 2023.

\bibitem{chen2023rank1}
X.~Chen, J.~Pan, B.~Li, and Q.~Wang.
\newblock An evolutionary algorithm based on rank-1 approximation for sparse
  large-scale multi-objective problems.
\newblock {\em Soft Computing}, 27:15853--15871, 2023.

\bibitem{zitzler2001spea2}
E.~Zitzler, M.~Laumanns, and L.~Thiele.
\newblock Spea2: Improving the strength pareto evolutionary algorithm.
\newblock Number 103, 2001.

\bibitem{deb1995simulated}
K.~Deb and R.~B. Agrawal.
\newblock Simulated binary crossover for continuous search space.
\newblock {\em Complex Systems}, 9(2):115--148, 1995.

\bibitem{deb1996combined}
K.~Deb and M.~Goyal.
\newblock A combined genetic adaptive search (geneas) for engineering design.
\newblock {\em Computer Science and Informatics}, 26:30--45, 1996.

\bibitem{su2022comparing}
Y.~Su, Z.~Jin, Y.~Tian, X.~Zhang, and K.~C. Tan.
\newblock Comparing the performance of evolutionary algorithms for sparse
  multi-objective optimization via a comprehensive indicator.
\newblock {\em IEEE Computational Intelligence Magazine}, 17(3):34--53, 2022.

\bibitem{kropp2023improved}
I.~Kropp, A.~P. Nejadhashemi, and K.~Deb.
\newblock Improved evolutionary operators for sparse large-scale multiobjective
  optimization problems.
\newblock {\em IEEE Trans. on Evolutionary Computation}, 2023.

\bibitem{qi2024two}
S.~Qi, R.~Wang, T.~Zhang, X.~Yang, R.~Sun, and L.~Wang.
\newblock A two-layer encoding learning swarm optimizer based on frequent
  itemsets for sparse large-scale multi-objective optimization.
\newblock {\em IEEE-CAA J. Automatica Sinica}, 11(6):1342--1357, 2024.

\bibitem{tian2023practical}
Y.~Tian, W.~Zhu, X.~Zhang, and Y.~Jin.
\newblock A practical tutorial on solving optimization problems via platemo.
\newblock {\em Neurocomputing}, 518:190--205, 2023.

\bibitem{bosman2003balance}
P.~A.~N. Bosman and D.~Thierens.
\newblock The balance between proximity and diversity in multiobjective
  evolutionary algorithms.
\newblock {\em IEEE Trans. on Evolutionary Computation}, 7(2):174--188, 2003.

\bibitem{zitzler1999multiobjective}
E.~Zitzler and L.~Thiele.
\newblock Multiobjective evolutionary algorithms: A comparative case study and
  the strength pareto approach.
\newblock {\em IEEE Trans. on Evolutionary Computation}, 3(4):257--271, 1999.

\bibitem{li2017how}
M.~Li, L.~Zhen, and X.~Yao.
\newblock How to read many-objective solution sets in parallel coordinates
  [educational forum].
\newblock {\em IEEE Computational Intelligence Magazine}, 12(4):88--100, 2017.

\end{thebibliography}
\end{document}